\documentclass{article}
\usepackage[accepted]{icml2021}

\usepackage[T1]{fontenc}
\usepackage{microtype}
\usepackage[hyperref, table]{xcolor}
\usepackage[hang, flushmargin]{footmisc}

\usepackage{math}
\usepackage{xkcd_color}

\usepackage[
  colorlinks=true,
  linkcolor={xkcd red},
  citecolor={xkcd blue},
  urlcolor={xkcd magenta},
  backref=section,
]{hyperref}
\usepackage[capitalise]{cleveref}

\usepackage{natbib}
\usepackage{graphicx}
\graphicspath{{./figures/}}

\usepackage{booktabs}
\usepackage{multirow}

\usepackage[inline]{enumitem}

\usepackage{amssymb}
\usepackage{siunitx}
\usepackage{amsthm}
\theoremstyle{definition}
\newtheorem{definition}{Definition}
\newtheorem{lemma}{Lemma}
\usepackage{thm-restate}

\newcommand{\Conv}{{\mathrm{Conv}}}
\newcommand{\TV}{{\mathrm{TV}}}
\newcommand{\KL}{{\mathrm{KL}}}
\newcommand{\Yt}{{\widetilde{Y}}}
\newcommand{\Rt}{{\widetilde{R}}}
\newcommand{\Lt}{{\widetilde{L}}}
\newcommand{\vpt}{{\widetilde{\vp}}}

\newcommand{\Yh}{{\widehat{Y}}}
\newcommand{\vph}{{\widehat{\vp}}}
\newcommand{\mTh}{{\widehat{\mT}}}

\begin{document}

\twocolumn[
\icmltitlerunning{%
Learning Noise Transition Matrix from Only Noisy Labels 
via Total Variation Regularization
}
\icmltitle{%
Learning Noise Transition Matrix from Only Noisy Labels\\
via Total Variation Regularization
}

\begin{icmlauthorlist}
\icmlauthor{Yivan Zhang}{utokyo,riken}
\icmlauthor{Gang Niu}{riken}
\icmlauthor{Masashi Sugiyama}{riken,utokyo}
\end{icmlauthorlist}

\icmlaffiliation{utokyo}{The University of Tokyo, Japan}
\icmlaffiliation{riken}{RIKEN AIP, Japan}

\icmlcorrespondingauthor%
{Yivan Zhang}{yivanzhang@ms.k.u-tokyo.ac.jp}

\icmlkeywords{%
Weakly-Supervised Learning,
Classification,
Label Noise,
Regularization,
Total Variation,
}

\vskip 0.3in
] 
\printAffiliationsAndNotice{}


\begin{abstract}
Many weakly supervised classification methods employ a \emph{noise transition matrix} to capture the class-conditional label corruption.
To estimate the transition matrix from noisy data, existing methods often need to estimate the noisy class-posterior, which could be unreliable due to the overconfidence of neural networks.
In this work, we propose a theoretically grounded method that can estimate the noise transition matrix and learn a classifier simultaneously, without relying on the error-prone noisy class-posterior estimation.
Concretely, inspired by the characteristics of the stochastic label corruption process, we propose \emph{total variation regularization}, which encourages the predicted probabilities to be more distinguishable from each other.
Under mild assumptions, the proposed method yields a \emph{consistent estimator} of the transition matrix.
We show the effectiveness of the proposed method through experiments on benchmark and real-world datasets.
\end{abstract}

\section{Introduction}
\label{sec:introduction}
Can we learn a correct classifier based on possibly incorrect examples?
The study of classification in the presence of \emph{label noise} has been of interest for decades \citep{angluin1988learning} and is becoming more important in the era of deep learning \citep{goodfellow2016deep}.
This issue can be caused by the use of imperfect surrogates of clean labels produced by annotation techniques for large-scale datasets such as crowdsourcing and web crawling \citep{fergus2005learning, jiang2020beyond}.
Unfortunately, without proper regularization, deep models could be more vulnerable to overfitting the label noise in the training data \citep{arpit2017closer, zhang2017understanding}, which affects the classification performance adversely.


\paragraph{Background.}

Early studies on learning from noisy labels can be traced back to the \emph{random classification noise} (RCN) model for binary classification \citep{angluin1988learning, long2010random, van2015learning}.
Then, RCN has been extended to the case where the noise rate depends on the classes, called the \emph{class-conditional noise} (CCN) model \citep{natarajan2013learning}.
The multiclass case is of central interest in recent years \citep{patrini2017making, goldberger2017training, han2018masking, xia2019anchor, yao2020dual}, where multiclass labels $Y$ flip into other categories $\Yt$ according to probabilities described by a fixed but unknown \emph{noise transition matrix} $\mT$, where $\mT_{ij} = p(\Yt = j | Y = i)$.
In this work, we focus on the multiclass CCN model.
Other noise models are discussed in \cref{app:ccn}.


\vspace{-1em}
\paragraph{Methodology.}

The unknown noise transition matrix in CCN has become a hurdle.
In this work, we focus on a line of research that aims to estimate the transition matrix from noisy data.
With a consistently estimated transition matrix, consistent estimation of the clean class-posterior is possible \citep{patrini2017making}.
To estimate the transition matrix, earlier work mainly relies on a given set of \emph{anchor points} \citep{liu2015classification, patrini2017making, yu2018learning}, i.e., instances belonging to the true class deterministically.
With anchor points, the transition matrix becomes identifiable based on the noisy class-posterior.
Further, recent work has attempted to detect anchor points in noisy data to mitigate the lack of anchor points in real-world settings \citep{xia2019anchor, yao2020dual}.

Nevertheless, even with a given anchor point set, these two-step methods of first estimating the transition matrix and then using it in neural network training face an inevitable problem --- the estimation of the noisy class-posterior.
The estimation error could be high due to the overconfidence of neural networks \citep{guo2017calibration, hein2019relu, rahimi2020intra} (see also \cref{app:overconfidence}).

In this work, we present an alternative methodology that does not rely on the error-prone estimation of the noisy class-posterior.
The key idea is as follows:
Note the fact that the noisy class-posterior vector $\vpt$ is given by the product of the noise transition matrix $\mT$ and the clean class-posterior vector $\vp$: $\vpt = \mT\T\vp$.
However, in the reverse direction, the decomposition of the product is not always unique, so $\mT$ and $\vp$ are not \emph{identifiable} from $\vpt$.
Thus, without additional assumption, existing methods for estimating $\mT$ from $\vpt$ could be unreliable.
However, if $\vp$ has some characteristics, e.g., $\vp$ is the ``cleanest'' among all the possibilities, then $\mT$ and $\vp$ become identifiable and it is possible to construct consistent estimators.
Concretely, we assume that \emph{anchor points exist in the dataset} to guarantee the identifiability, but we do not need to explicitly model or detect them (\cref{thm:decomposition}).

Further, note that the mapping $\vp \mapsto \mT\T\vp$ is a \emph{contraction} over the probability simplex relative to the \emph{total variation distance} \citep{del2003contraction}.
That is, the ``cleanest'' $\vp$ has the property that pairs of $\vp$ are more distinguishable from each other.
Based on this motivation, we propose \emph{total variation regularization} to find the ``cleanest'' $\vp$ and consistently estimate $\mT$ simultaneously (\cref{thm:consistency}).


\begin{figure}
\centering
\includegraphics[width=\linewidth]
{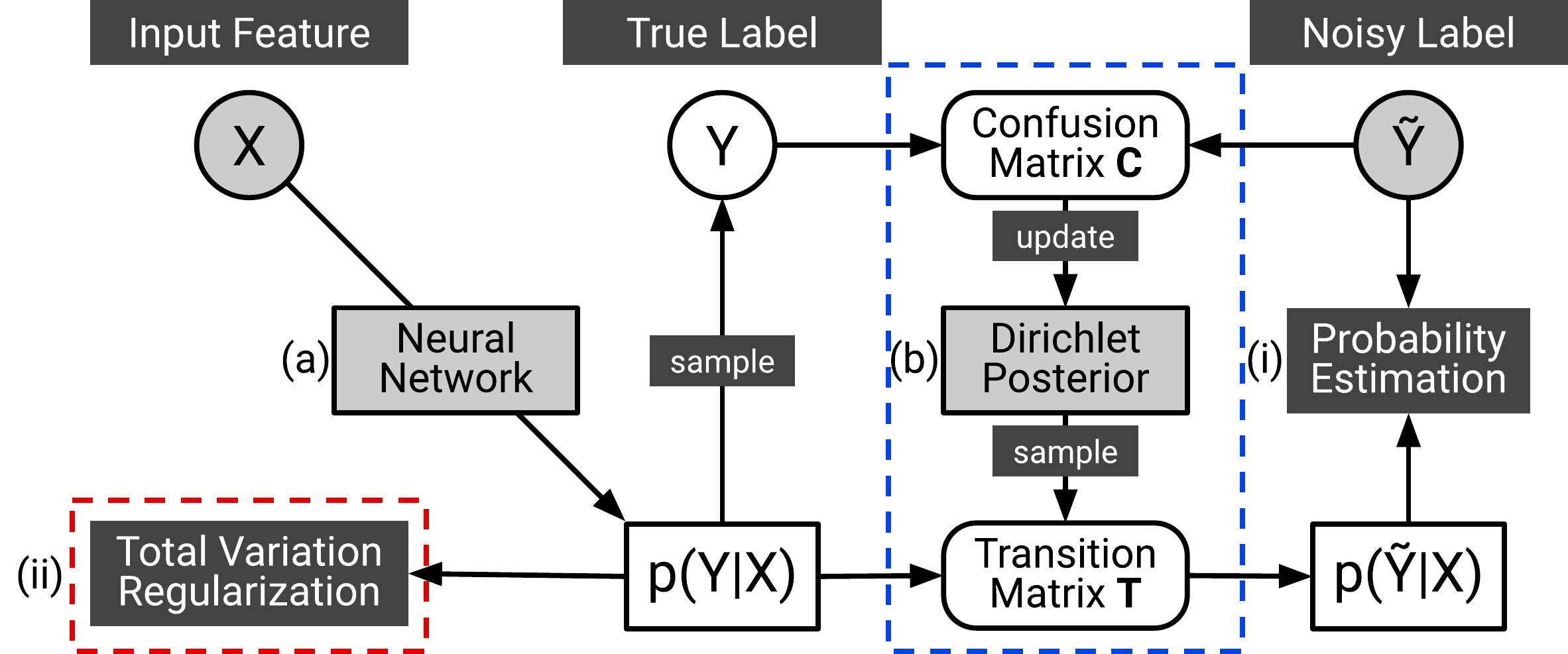}
\vspace{-2em}
\caption{%
An illustration of the \textbf{proposed method}.
Our model has two modules: 
(a) a neural network for predicting $\vp(Y | X)$; and
(b) a \textcolor{xkcd blue}{\emph{Dirichlet posterior}} for the noise transition matrix $\mT$, whose concentration parameters are updated using the confusion matrix obtained during training.
The learning objective in \cref{eq:learning_objective} also contains two parts:
(i) the usual cross entropy loss for classification from noisy labels in \cref{eq:nll}; and
(ii) a \textcolor{xkcd red}{\emph{total variation regularization}} term for the predicted probability in \cref{eq:regularization}.
}
\label{fig:model}
\end{figure}


\vspace{-1em}
\paragraph{Our contribution.}

In this paper, we study the class-conditional noise (CCN) problem and propose a method that can estimate the noise transition matrix and learn a classifier simultaneously, given only noisy data.
The key idea is to regularize the predicted probabilities to be more distinguishable from each other using the pairwise total variation distance.
Under mild conditions, the transition matrix becomes identifiable and a consistent estimator can be constructed.

Specifically, we study the characteristics of the class-conditional label corruption process and construct a partial order within the equivalence class of transition matrices in terms of the total variation distance in \cref{sec:motivation}, which motivates our proposed method.
In \cref{ssec:regularization}, we present the proposed \emph{total variation regularization} and the theorem of consistency (\cref{thm:consistency}).
In \cref{ssec:matrix}, we propose a conceptually novel method based on \emph{Dirichlet distributions} for estimating the transition matrix.
Overall, the proposed method is illustrated in \cref{fig:model}.

\begin{figure*}
\centering
\includegraphics[width=\linewidth]
{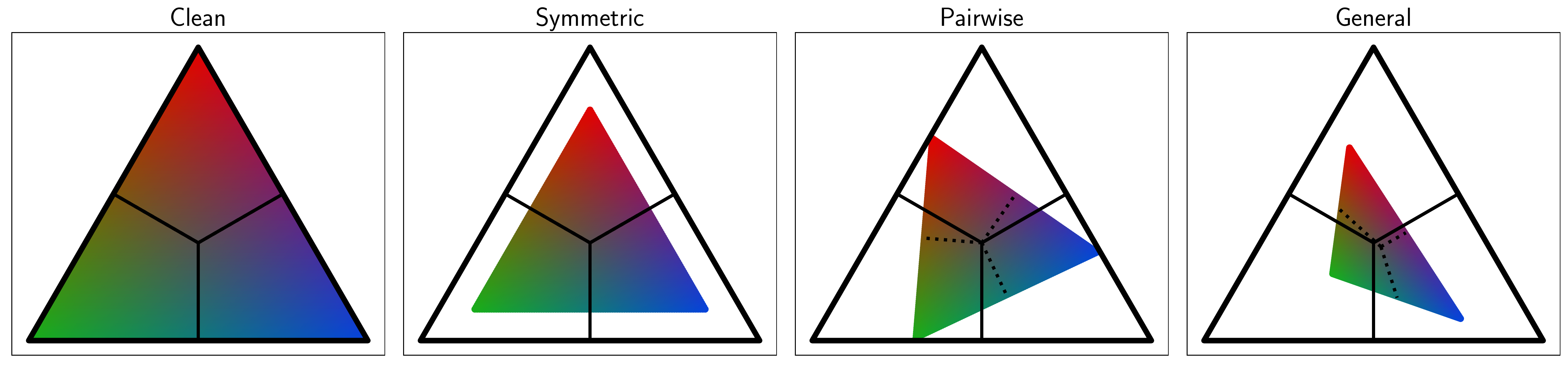}
\vspace{-2em}
\caption{%
Illustrations of the \textbf{noise transition matrix} as a linear transformation $\Delta \to \Conv(\mT)$ when $K=3$, including symmetric, pair flipping, and general noises.
The outer black triangle is the simplex $\Delta$ and the inner colored triangle is the convex hull $\Conv(\mT)$.
Solid lines are the decision boundaries of argmax and dotted lines are the ones after the transformation.
Note that different transition matrices affect the decision boundary differently. 
Adding symmetric noise does not change the decision boundary.
}
\label{fig:noise}
\end{figure*}

\section{Problem: Class-Conditional Noise}
\label{sec:problem}
In this section, we review the notation, assumption, and related work in learning with \emph{class-conditional noise} (CCN).


\subsection{Noisy Labels}

Let $X \in \sX$ be the \emph{input feature}, and $Y \in \{1, \dots, K\}$ the \emph{true label}, where $K$ is the number of classes.
In fully supervised learning, we can fit a discriminative model for the conditional probability $p(Y | X)$ using an i.i.d.~sample of $(X, Y)$-pairs.
However, observed labels may be corrupted in many real-world applications.
Treating the corrupted label as correct usually leads to poor performance \citep{arpit2017closer, zhang2017understanding}.
In this work, we explicitly model this label corruption process.
We denote the \emph{noisy label} by $\Yt \in \{1, \dots, K\}$.
The goal is to predict $Y$ from $X$ based on an i.i.d.~sample of $(X, \Yt)$-pairs.


\subsection{Class-Conditional Noise (CCN)}
\label{ssec:ccn}

Next, we formulate the \emph{class-conditional noise} (CCN) model \citep{natarajan2013learning, patrini2017making}.

In CCN, we have the following assumption:
\begin{equation}
\label{eq:ccn}
p(\Yt | Y, X) = p(\Yt | Y)
.
\end{equation}
That is, the noisy label $\Yt$ only depends on the true label $Y$ but not on $X$.
Then, we can relate the \emph{noisy class-posterior} $p(\Yt | X)$ and the \emph{clean class-posterior} $p(Y | X)$ by
\begin{equation}
\label{eq:ccn_sum}
\textstyle
  p(\Yt | X) = \sum_{Y=1}^K p(\Yt | Y) p(Y | X)
.
\end{equation}
Note that the clean class-posterior $p(Y | X)$ can be seen as a vector-valued function
$\vp(Y | X=x): \sX \to \Delta^{K-1}
\defeq [p(Y=1 | X=x), \ldots, p(Y=K | X=x)]\T$,
and so can the noisy class-posterior $p(\Yt | X)$.%
\footnote{%
$\Delta^{K-1}$ denotes the $(K-1)$-dimensional probability simplex.\\
The superscript in $\Delta^{K-1}$ is omitted hereafter.
}
Also, $p(\Yt | Y)$ can be written in matrix form:
$\mT \in \sT \subset [0, 1]^{K \times K}$ 
with elements $\mT_{ij} = p(\Yt = j | Y = i)$ 
for $i, j \in \{1, \ldots, K\}$,
where $\sT$ is the set of all full-rank row stochastic matrices.
Here, $\mT$ is called a \emph{noise transition matrix}.
Then, with the vector and matrix notation, \cref{eq:ccn_sum} can be rewritten as
\begin{equation}
\label{eq:ccn_vec}
  \vp(\Yt | X) = \mT\T \vp(Y | X)
.
\end{equation}
Note that multiplying $\mT$ is a linear transformation from the simplex $\Delta$ to the convex hull $\Conv(\mT)$ induced by rows of $\mT$,%
\footnote{%
Here, $\Conv(\mT)$ is a shorthand for the \emph{convex hull} of the set of vectors $\{\mT_i | i=1, \ldots, K\}$ within the simplex $\Delta$.
}
which is illustrated in \cref{fig:noise}.

In the context of learning from noisy labels, we further assume $\mT$ to be \emph{diagonally dominant} in the sense that $\mT_{ii} > \mT_{ij}$ for $i \neq j$.
Although this formulation can be also used for learning from \emph{complementary labels}, where $\mT_{ii} = 0$ or $\mT_{ii} < \mT_{ij}$ for $i \neq j$ \citep{ishida2017learning, yu2018learning}.


\subsection{Learning with Known Noise Transition Matrix}
\label{ssec:known_T}

If the ground-truth noise transition matrix $\mT$ is known, $\vp(Y | X)$ is \emph{identifiable} based on observations of $\vp(\Yt | X)$, which means that different $\vp(Y | X)$ must generate distinct $\vp(\Yt | X)$.
Therefore, we can consistently recover $\vp(Y | X)$ by estimating $\vp(\Yt | X)$ using the relation in \cref{eq:ccn_vec} \citep{patrini2017making, yu2018learning}.
However, if $\mT$ and $\vp(Y | X)$ are both unknown, then they are both \emph{partially identifiable} because there could be multiple observationally equivalent $\mT$ and $\vp(Y | X)$ whose product equals $\vp(\Yt | X)$.
Thus, it is impossible to estimate both $\mT$ and $\vp(Y | X)$ from $\vp(\Yt | X)$ without any additional assumption.

Concretely, let $\vph(Y | X; W)$ parameterized by $W \in \sW$ be a differentiable model for the true label,%
\footnote{%
$W$ is sometimes omitted to keep the notation uncluttered.
}
and $\mTh \in \sT$ be an estimator for the noise transition matrix.
We consider a sufficiently large function class of $\vph(Y | X; W)$ that contains the ground-truth $\vp(Y | X)$, i.e., 
$
\exists W^* \in \sW, 
\vph(Y | X; W^*) = \vp(Y | X) 
\almosteverywhere
$
In practice, we use an expressive deep neural network \citep{goodfellow2016deep} as $\vph(Y | X; W)$.

Then, let us consider the \emph{expected Kullback–Leibler (KL) divergence} as the learning objective:
\begin{equation}
\label{eq:kl}
  L_0(W, \mTh)
\defeq
  \E_{X \sim p(X)}
  \brackets*{
  D_\KL \diver*{\vp(\Yt | X)}{\mTh\T\vph(Y | X; W)}
  }
,
\end{equation}
which is related to the \emph{expected negative log-likelihood} or the \emph{cross-entropy loss} in the following way:
\begin{align}
\label{eq:nll}
  L(W, \mTh)
&\defeq
  \E_{X, \Yt \sim p(X, \Yt)}\brackets*{
  -\log (\mTh_{\Yt}\T \vph(Y | X; W))}
\\
&= 
  L_0(W, \mTh) + \Eta(\Yt | X)
,
\end{align}
where the second term $\Eta(\Yt | X)$ is the conditional entropy of $\Yt$ given $X$, which is a constant w.r.t.~$W$ and $\mTh$.
Note that $L(W, \mTh)$ is minimized if and only if 
$L_0(W, \mTh) = 0$.

\newpage
When $L(W, \mTh)$ is empirically estimated and optimized based on a finite sample of $(X, \Yt)$-pairs, we can ensure that 
$\mTh\T \vph(Y | X) \converged \mT\T \vp(Y | X)$
as the sample size increases, but we can not guarantee that
$\vph(Y | X) \converged \vp(Y | X)$ 
due to non-identifiability.%
\footnote{%
$\converged$ denotes the convergence in distribution.
}
The latter holds only when the ground-truth $\mT$ is used as $\mTh$ \citep{patrini2017making}.


\subsection{Learning with Unknown Noise Transition Matrix}
\label{ssec:unknown_T}

In real-world applications, the ground-truth $\mT$ is usually unknown.
Some existing two-step methods attempted to transform this problem to the previously solved one by first estimating the noise transition matrix and then using it in neural network training.
Since it is rare to have both clean labels $Y$ and noisy labels $\Yt$ for the same instance $X$, several methods are proposed to estimate $\mT$ from only noisy data.

Existing methods usually rely on a separate set of \emph{anchor points} \citep{liu2015classification, patrini2017making, yu2018learning, xia2019anchor, yao2020dual}, which are defined as follows:
\begin{definition}[Anchor point]
\label{def:anchor}
An instance $x$ is called an anchor point for class $i$ if
$p(Y = i | X = x) = 1$.
\end{definition}
Based on an anchor point $x$ for class $i$, we have
\begin{equation}
  \vp(\Yt | X = x) 
= \mT\T \vp(Y | X = x)
= \mT_i
.
\end{equation}
Thus, we can first estimate $\vp(\Yt | X)$ and then calculate the value on anchor points to obtain an estimate of $\mT$.
However, if we cannot find such anchor points in real-world datasets easily, the aforementioned method can not be applied.

A workaround is to detect anchor points from all noisy data, assuming that they exist in the dataset.
Further revision of the transition matrix before \citep{yao2020dual} or during \citep{xia2019anchor} the second stage of training can be adopted to improve the performance.


\begin{figure*}
\centering
\includegraphics[width=\linewidth]
{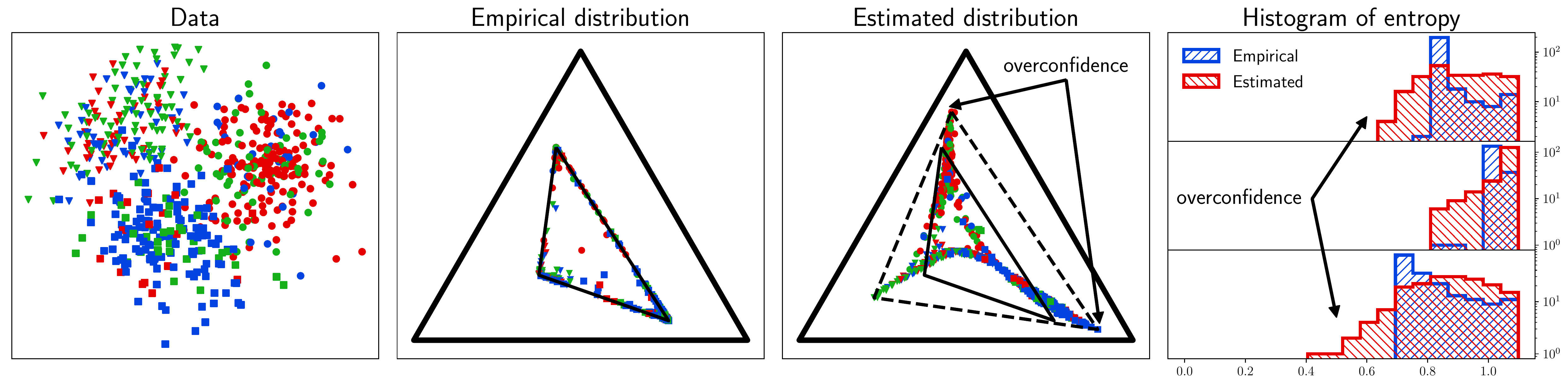}
\vspace{-2em}
\caption[]{%
An example of \textbf{overconfident predictions} yield from neural networks.
Notation:
\begin{itemize*}[itemjoin={{; }}]
\item Shape: true labels
\item Color: observed labels
\item Location in the simplex: ground-truth/estimated probabilities
\item Solid triangle: convex hull $\Conv(\mT)$
\item Dashed triangle: convex hull of an estimated transition matrix
\item Vertices of the triangle: anchor points
\end{itemize*}.
Without knowing $\mT$ and the constraint that $\vp(\Yt | X)$ should be within $\Conv(\mT)$, a neural network trained with noisy labels tends to output high-confident (low-entropy) predictions outside of $\Conv(\mT)$.
Therefore, $\mT$ may be poorly estimated based on the overconfident noisy class-posterior.
}
\label{fig:overconfidence}
\end{figure*}


Nevertheless, such two-step methods based on anchor points face an inevitable problem --- the estimation of the noisy class-posterior $\vp(\Yt | X)$ using possibly over-parameterized neural networks trained with noisy labels.
We point out that the estimation error could be high in this step because of the overconfidence problem of deep neural networks \citep{guo2017calibration, hein2019relu}.
If no revision is made, errors in the first stage may lead to major errors in the second stage.

\Cref{fig:overconfidence} illustrates an example of the overconfidence.
As discussed in \cref{ssec:ccn}, $\vp(\Yt | X)$ should be within the convex hull $\Conv(\mT)$.
However, without knowing $\mT$ and this constraint, a neural network trained with noisy labels tends to output overconfident probabilities that are outside of $\Conv(\mT)$.
Note that over-parameterized neural networks trained with clean labels could also be overconfident and several re-calibration methods are developed to alleviate this issue \citep{guo2017calibration, kull2019beyond, hein2019relu, rahimi2020intra}.
However, in \cref{app:overconfidence} we demonstrate that estimating the noise class-posterior causes a significantly worse overconfidence issue than estimating the clean one.
Consequently, transition matrix estimation may suffer from poorly estimated noisy class-posteriors, which leads to performance degradation.

In contrast to existing methods, our proposed method only uses the product $\mTh\T\vph(Y | X)$ as an estimate of $\vp(\Yt | X)$ and never estimates $\vp(\Yt | X)$ directly using neural networks.

\section{Motivation}
\label{sec:motivation}
In this section, we take a closer look at the class-conditional label corruption process and construct an equivalence class and a partial order for the noise transition matrix, which motivates our proposed method.
Concretely, we show that the \emph{contraction property} of the stochastic matrices leads to a partial order of the transition matrices, which can be used to find the ``cleanest'' clean class-posterior.


\subsection{Transition Matrix Equivalence}

Recall that $\sT$ is the set of full-rank row stochastic matrices, which is \emph{closed under multiplication}.
Based on this, we first define an \emph{equivalence relation} of an ordered pair of transition matrices induced by the product:
\begin{definition}[Transition matrix equivalence]
\label{def:equivalence}
\begin{equation*}
(\mU, \mV) \sim (\mU', \mV')
\Leftrightarrow 
\mU\mV = \mU'\mV'
.
\end{equation*}
\end{definition}
\vspace{-1em}
The corresponding equivalence class with a product $\mW$ is denoted by $[\mW]$.
Specially, for the identity matrix $\mI$, $[\mI]$ contains pairs of permutation matrices $(\mP, \mP\inv)$; for a non-identity matrix $\mW$, $[\mW]$ contains at least two distinct elements $(\mW, \mI)$ and $(\mI, \mW)$ and possibly infinitely many other elements.

Now, consider the equivalence class $[\mT]$ for the ground-truth noise transition matrix $\mT$ in our problem.
Then, any element $(\mU, \mV) \in [\mT]$ corresponds to a possible optimal solution of \cref{eq:nll}: 
$\mTh = \mV$ and $\vph(Y | X; W) = \mU\T \vp(Y | X)$, 
given that such a parameter $W$ exists.
Among possibly infinitely many possibilities, only $(\mI, \mT)$ is of our central interest.
However, it is possible to get infinitely many other wrong ones, such as $(\mT, \mI)$, which corresponds to a model that predicts the noisy class-posterior and chooses the transition matrix to be the identity matrix $\mI$.


\subsection{Transition Matrix Decomposition}

Next, consider the reverse direction:
if we obtain an optimal solution of \cref{eq:nll}, $\mTh$ and $\vph(Y | X)$, is there a $\mU$ such that $\vph(Y | X) = \mU\T \vp(Y | X)$?
The answer is yes if there are anchor points for each class in the dataset, which can be proved using the following theorem:
\begin{restatable}[Transition matrix decomposition]{theorem}{decomposition}
\label{thm:decomposition}
For two row stochastic matrices $\mW, \mV \in \sT$,
if $\forall \vp \in \Delta$, $\exists \vq \in \Delta$,
s.t.~$\mW\T\vp = \mV\T\vq$,
then $\exists$ a row stochastic matrix $\mU \in \sT$,
s.t.~$\mW = \mU\mV$ 
and $\forall \vp \in \Delta$, $\vq = \mU\T\vp$.
\vspace{-1em}
\begin{proof}
Let $\vp$ be $\ve_i$ and denote the corresponding $\vq$ by $\vq_i$ for $i = 1, \dots, K$. 
Then $\mU = [\vq_1, \dots, \vq_K]\T$.
\end{proof}
\end{restatable}
Here, $\ve_i$ is the $i$-th standard basis and $\vp(Y | X = x) = \ve_i$ means that $x$ is an anchor point for the class $i$.
Consequently, we can derive that if there are anchor points for each class in the dataset, given an estimated transition matrix $\mTh$ and an estimated clean class-posterior $\vph(Y | X)$ from an optimal solution of \cref{eq:nll}, we know that there is an \emph{implicit} row stochastic matrix $\mU$ such that $\vph(Y | X) = \mU\T \vp(Y | X)$.
In other words, the estimate $\vph(Y | X)$ may still contain class-conditional label noise, which is described by $\mU$.

We point out that the existence of anchor points is a \emph{sufficient but not necessary} condition for the existence of the $\mU$ above.
If anchor points do not exist, we may or may not find such a $\mU$.
Also note that we will not try to detect anchor points from noisy data.

More importantly, we have no intention of estimating $\mU$ explicitly.
In this work, we only use the fact that there is a one-to-one correspondence between optimal solutions of \cref{eq:nll} and elements in the equivalence class $[\mT]$ under the above assumption.
Based on this fact, we can study the equivalence class $[\mT]$ or the properties of the implicit $\mU$ instead, which is easier to deal with.


\subsection{Transition Matrix as a Contraction Mapping}

Next, we attempt to \emph{break the equivalence} introduced above by examining the characteristics of this consecutive class-conditional label corruption process.

We start with the definition of the \emph{total variation distance} $d_\TV(\cdot, \cdot)$ between pairs of categorical probabilities:
\begin{equation}
\label{eq:tv}
d_\TV(\vp, \vq) 
\defeq 
\frac12 \norm{\vp - \vq}_1
,
\end{equation}
where $\norm{\cdot}_1$ denotes the $\ell_1$ norm.
Then, from the theory of Markov chains, we know that the mapping $\Delta \to \Conv(\mU)$ defined by $\vp \mapsto \mU\T\vp$ is a \emph{contraction mapping} over the simplex $\Delta$ relative to the total variation distance \citep{del2003contraction}, which means that
$\forall \mU \in \sT, \forall \vp, \vq \in \Delta$,
\begin{equation}
\label{eq:contraction}
d_\TV(\mU\T\vp, \mU\T\vq) \leq d_\TV(\vp, \vq)
.
\end{equation}
%


\subsection{Transition Matrix Partial Order}

Finally, based on this contraction property of the stochastic matrices, we can introduce a \emph{partial order} induced by the total variation distance within the equivalence class $[\mT]$:
\begin{definition}[Transition matrix partial order]
\label{def:partial_order}
\begin{gather*}
(\mU, \mV) \preceq (\mU', \mV') 
\Leftrightarrow
\\
\forall \vp, \vq \in \Delta,
d_\TV(\mU\T\vp, \mU\T\vq) \leq d_\TV(\mU^{\prime\mathsf{T}}\vp, \mU^{\prime\mathsf{T}}\vq)
.
\end{gather*}
\end{definition}
Note that $(\mI, \mT)$ is the \emph{unique greatest element} because of \cref{eq:contraction}.
Despite the fact that there could be incomparable elements, we may gradually increase the total variation to find $(\mI, \mT)$.
Then, with the help of this partial order, it is possible to estimate both $\vp(Y | X)$ and $\mT$ simultaneously, which is discussed in the following section.

\section{Proposed Method}
\label{sec:method}
In this section, we present our proposed method.
Overall, the proposed method is illustrated in \cref{fig:model}.

Summarizing our motivation discussed in \cref{sec:motivation}, we found that if anchor points exist in the dataset, estimating both the transition matrix $\mT$ and the clean class-posterior $\vp(Y | X)$ by training with the cross-entropy loss in \cref{eq:nll} results in a solution in the form $\vph(Y | X) = \mU\T \vp(Y | X)$, where $\mU$ is an unknown transition matrix (\cref{thm:decomposition}).
Then, we pointed out that the stochastic matrices have the contraction property shown in \cref{eq:contraction} so that the ``cleanest'' clean class-posterior has the highest pairwise total variation defined in \cref{eq:tv}.
Based on this fact, we can regularize the predicted probabilities to be more distinguishable from each other to find the optimal solution, as discussed below.


\subsection{Total Variation Regularization}
\label{ssec:regularization}

First, 
we discuss how to enforce our preference of more distinguishable predictions in terms of the total variation distance.
We start with defining the \emph{expected pairwise total variation distance}:
\begin{equation}
\label{eq:regularization}
\begin{aligned}
  R(W) 
&\defeq 
  \E_{x_1 \sim p(X)}\E_{x_2 \sim p(X)}
  [d_\TV(\vph_1, \vph_2)]
,
\\
\where
  \vph_i &\defeq \vph(Y | X = x_i; W), \quad i = 1, 2
.
\end{aligned}
\end{equation}
Note that this \emph{data-dependent} term depends on $X$ but not on $Y$ nor on $\Yt$.

Then, we adopt the learning objective in the KL-divergence form in \cref{eq:kl}, combine it with the expected pairwise total variation distance in \cref{eq:regularization}, and formulate our approach in the form of \emph{constrained optimization}, as stated in the following theorem:
\begin{restatable}[Consistency]{theorem}{consistency}
\label{thm:consistency}
Given a finite i.i.d.~sample of $(X,\Yt)$-pairs of size $N$, where anchor points (\cref{def:anchor}) for each class exist in the sample, let $\Lt_0(W, \mTh)$ and $\Rt(W)$ be the empirical estimates of $L_0(W, \mTh)$ in \cref{eq:kl} and $R(W)$ in \cref{eq:regularization}, respectively.
Assume that the parameter space $\sW$ is compact.
Let $(W^\circ, \mTh^\circ)$ be an optimal solution of the following constrained optimization problem:
\begin{equation}
\label{eq:constrained_optimization}
\max_W \Rt(W) \st \Lt_0(W, \mTh) = 0
.
\end{equation}
Then, $\mTh^\circ$ is a consistent estimator of the transition matrix $\mT$; 
and $\vph(Y | X; W^\circ) \converged \vp(Y | X) \almosteverywhere$ as $N \to \infty$.
\end{restatable}
The proof is given in \cref{app:proof}.
Informally, we make use of \cref{thm:decomposition}, the property of the KL-divergence, and the contraction property of the transition matrix.

In practice, the constrained optimization in \cref{eq:constrained_optimization} can be solved via the following Lagrangian \citep{kuhn1951nonlinear}:
\begin{equation}
\label{eq:learning_objective}
\mathcal{L}(W, \mTh) \defeq \Lt_0(W, \mTh) - \gamma \Rt(W)
,
\end{equation}
where $\gamma \in \R_{> 0}$ is a parameter controlling the importance of the regularization term.
We call such a regularization term a \emph{total variation regularization}.
This Lagrangian technique has been widely used in the literature \citep{cortes1995support, kloft2009efficient, higgins2017beta, li2021provably}.
When the total variation regularization term is empirically estimated and optimized, we can sample a fixed number of pairs to reduce the additional computational cost.


\subsection{Transition Matrix Estimation}
\label{ssec:matrix}

Next, we discuss the estimation of the transition matrix $\mT$.
In contrast to existing methods \citep{patrini2017making, xia2019anchor, yao2020dual}, we adopt a one-step training procedure to obtain both $\vph(Y | X)$ and $\mTh$ simultaneously.

\paragraph{Gradient-based estimation.}

First, note that the learning objective \cref{eq:learning_objective} is differentiable w.r.t.~$\mTh$.
As a baseline, it is sufficient to use gradient-based optimization for $\mTh$.
In practice, we apply softmax to an unconstrained matrix in $\R^{K \times K}$ to ensure that $\mTh \in \sT$.
Then, $\mTh$ is estimated by optimizing $\mathcal{L}(W, \mTh)$ using stochastic gradient descent (SGD) or its variants (e.g., \citet{kingma2014adam}).

\paragraph{Dirichlet posterior update.}

The additional total variation regularization term \cref{eq:regularization} is irrelevant to $\mTh$ so we are free to use other optimization methods besides gradient-based methods.
To capture the uncertainty of the estimation of $\mT$ during different stages of training, we propose an alternative \emph{derivative-free} approach that uses Dirichlet distributions to model $\mT$.
Concretely, let the posterior of $\mT$ be
\begin{equation}
\widehat{\mT}_i
\sim \Dirichlet(\mAlpha_i)
\quad (i = 1, \dots, K)
,
\end{equation}
where $\mAlpha_i \in \R_{>0}^K$ is the concentration parameter.
Denote the \emph{confusion matrix} by $\mC \in \N_{\geq 0}^{K \times K}$, where its element $\mC_{ij}$ is the number of instances that are predicted to be $\Yh = i$ via sampling $\Yh \sim \vph(Y | X; W)$ but are labeled as $\Yt = j$ in the noisy dataset.
In other words, we use a posterior of $p(\Yt | \Yh)$ to approximate $p(\Yt | Y)$ during training.

Then, inspired by the closed-form posterior update rule for the Dirichlet-multinomial conjugate \citep{diaconis1979conjugate}:
\begin{equation}
  \mAlpha^{(\mathrm{posterior})}
= \mAlpha^{(\mathrm{prior})} 
+ \mC^{(\mathrm{observation})}
,
\end{equation}
we update the concentration parameters $\mAlpha$ during training using the confusion matrix $\mC$ via the following update rule:
\begin{equation}
\label{eq:update}
  \mAlpha 
\leftarrow
  \beta_1 \mAlpha + \beta_2 \mC
,
\end{equation}
where $\vbeta = (\beta_1, \beta_2)$ are fixed hyperparameters that control the convergence of $\mAlpha$.
We initialize $\mAlpha$ with an appropriate diagonally dominant matrix to reflect our prior knowledge of noisy labels.
For each batch of data, we sample a noise transition matrix $\widehat{\mT}$ from the Dirichlet posterior and use it in our learning objective in \cref{eq:learning_objective}.

The idea is that the confusion matrix $\mC$ at any stage during training is a crude estimator of the true noise transition matrix $\mT$, then we can improve this estimator based on information obtained during training.
Because at earlier stage of training, this estimator is very crude and may be deviated from the true one significantly, we use a decaying factor $\beta_1$ close to $1$ (e.g., $0.999$) to let the model gradually ``forget'' earlier information.
Meanwhile, $\beta_2$ controls the variance of the Dirichlet posterior during training, which is related to the learning rate and batch size.
At early stages, the variance is high so the model is free to explore various transition matrices; as the model converges, the estimation of the transition matrix also becomes more precise so the posterior would concentrate around the true one.

\begin{table*}[t]
\centering
\caption{%
\textbf{Accuracy} ($\%$) on the MNIST, CIFAR-10, and CIFAR-100 datasets.
We reported ``mean (standard deviation)'' of $10$ trials.}
\label{tab:accuracy}
\begin{tabular}{cccccccc}
\toprule
&& (a) Clean & (b) Symm. & (c) Pair & (d) Pair$^2$ & (e) Trid. & (f) Rand.
\\

\midrule
\multirow{8}{*}{\rotatebox[origin=c]{90}{MNIST}}
& MAE & $98.72(0.09)$ & $98.00(0.14)$ & $91.46(7.40)$ & $89.79(6.11)$ & $96.22(3.87)$ & $34.07(31.98)$
\\
& CCE & $\mathbf{99.21(0.04)}$ & $98.13(0.16)$ & $94.70(0.64)$ & $94.86(0.67)$ & $96.78(0.22)$ & $95.68(1.31)$
\\
& GCE & $99.12(0.06)$ & $98.41(0.12)$ & $93.79(1.04)$ & $94.06(0.63)$ & $96.60(0.14)$ & $96.28(0.93)$
\\
& Forward & $\mathbf{99.18(0.05)}$ & $98.00(0.24)$ & $94.37(1.00)$ & $94.84(0.53)$ & $96.54(0.29)$ & $95.95(1.49)$
\\
& T-Revision & $\mathbf{99.20(0.06)}$ & $98.01(0.14)$ & $94.19(0.78)$ & $95.24(0.74)$ & $96.76(0.15)$ & $96.62(0.70)$
\\
& Dual-T & $99.16(0.05)$ & $\mathbf{98.58(0.12)}$ & $\mathbf{99.06(0.07)}$ & $\mathbf{99.03(0.06)}$ & $\mathbf{99.04(0.05)}$ & $\mathbf{98.79(0.17)}$
\\
& TVG & $99.16(0.06)$ & $\mathbf{98.55(0.09)}$ & $94.26(0.59)$ & $95.42(0.44)$ & $97.78(0.56)$ & $97.67(0.84)$
\\
& TVD & $\mathbf{99.18(0.07)}$ & $\mathbf{98.56(0.08)}$ & $\mathbf{99.09(0.08)}$ & $\mathbf{99.00(0.07)}$ & $\mathbf{99.03(0.08)}$ & $\mathbf{98.82(0.11)}$
\\

\midrule
\multirow{8}{*}{\rotatebox[origin=c]{90}{CIFAR10}}
& MAE & $66.47(4.76)$ & $57.23(4.15)$ & $44.29(2.23)$ & $42.43(1.66)$ & $43.43(2.69)$ & $26.95(5.45)$
\\
& CCE & $\mathbf{91.87(0.19)}$ & $75.71(0.57)$ & $65.54(0.66)$ & $65.23(0.85)$ & $76.07(0.61)$ & $70.44(1.98)$
\\
& GCE & $89.25(0.17)$ & $\mathbf{83.68(0.29)}$ & $71.49(1.18)$ & $69.66(0.57)$ & $82.14(0.41)$ & $78.07(2.16)$
\\
& Forward & $\mathbf{91.87(0.15)}$ & $76.18(0.63)$ & $65.42(0.92)$ & $65.65(1.11)$ & $76.41(0.50)$ & $70.86(2.19)$
\\
& T-Revision & $91.72(0.18)$ & $75.51(0.59)$ & $65.49(0.97)$ & $65.70(0.66)$ & $76.18(0.80)$ & $71.22(1.62)$
\\
& Dual-T & $\mathbf{91.75(0.18)}$ & $82.85(0.42)$ & $80.86(1.03)$ & $79.61(1.20)$ & $\mathbf{88.11(0.28)}$ & $84.33(2.11)$
\\
& TVG & $91.61(0.14)$ & $82.60(0.38)$ & $\mathbf{89.78(0.16)}$ & $\mathbf{88.36(0.24)}$ & $\mathbf{88.07(0.25)}$ & $\mathbf{86.19(0.52)}$
\\
& TVD & $91.00(0.13)$ & $83.03(0.24)$ & $88.47(0.29)$ & $86.96(0.35)$ & $87.44(0.16)$ & $\mathbf{85.86(0.46)}$
\\

\midrule
\multirow{8}{*}{\rotatebox[origin=c]{90}{CIFAR100}}
& MAE & $11.23(1.02)$ & $7.89(0.67)$ & $6.94(1.11)$ & $6.60(0.74)$ & $7.45(0.55)$ & $7.15(0.98)$
\\
& CCE & $\mathbf{70.58(0.29)}$ & $42.94(0.47)$ & $44.00(0.71)$ & $41.37(0.27)$ & $46.55(0.54)$ & $42.41(0.48)$
\\
& GCE & $57.10(0.85)$ & $48.66(0.58)$ & $45.27(0.85)$ & $43.67(0.94)$ & $50.98(0.33)$ & $48.66(0.63)$
\\
& Forward & $\mathbf{70.58(0.28)}$ & $44.32(0.64)$ & $44.17(0.57)$ & $42.07(0.55)$ & $47.48(0.40)$ & $43.15(0.53)$
\\
& T-Revision & $\mathbf{70.47(0.26)}$ & $46.52(0.57)$ & $44.08(0.42)$ & $42.01(0.52)$ & $47.59(0.60)$ & $45.33(0.40)$
\\
& Dual-T & $\mathbf{70.56(0.28)}$ & $55.92(0.60)$ & $46.22(0.72)$ & $44.74(0.65)$ & $61.68(0.51)$ & $57.92(0.50)$
\\
& TVG & $70.02(0.30)$ & $\mathbf{57.33(0.42)}$ & $45.68(0.85)$ & $44.38(0.72)$ & $54.23(0.53)$ & $\mathbf{59.85(0.61)}$
\\
& TVD & $69.93(0.21)$ & $52.54(0.45)$ & $\mathbf{56.02(0.82)}$ & $\mathbf{49.18(0.53)}$ & $\mathbf{62.45(0.44)}$ & $53.95(0.47)$
\\

\bottomrule
\end{tabular}
\end{table*}

\begin{table*}[t]
\centering
\caption{%
\textbf{Average total variation} ($\times 100$) on the MNIST, CIFAR-10, and CIFAR-100 datasets.
We reported ``mean (standard deviation)'' of $10$ trials.
}
\label{tab:tv}
\begin{tabular}{cccccccc}
\toprule
&& (a) Clean & (b) Symm. & (c) Pair & (d) Pair$^2$ & (e) Trid. & (f) Rand.
\\

\midrule
\multirow{5}{*}{\rotatebox[origin=c]{90}{MNIST}}
& Forward & $\mathbf{0.00(0.00)}$ & $34.14(3.03)$ & $39.71(0.15)$ & $41.98(0.82)$ & $38.33(0.93)$ & $30.45(2.16)$
\\
& T-Revision & $0.03(0.02)$ & $32.94(3.22)$ & $39.87(0.08)$ & $41.50(0.50)$ & $38.39(1.34)$ & $29.35(1.85)$
\\
& Dual-T & $0.12(0.02)$ & $7.12(0.99)$ & $3.90(0.66)$ & $3.59(0.58)$ & $3.11(0.88)$ & $10.63(0.90)$
\\
& TVG & $2.36(0.01)$ & $\mathbf{1.47(0.13)}$ & $39.29(0.03)$ & $32.17(0.93)$ & $14.11(5.21)$ & $7.33(4.25)$
\\
& TVD & $2.06(0.12)$ & $1.96(0.17)$ & $\mathbf{2.12(0.21)}$ & $\mathbf{2.12(0.10)}$ & $\mathbf{1.92(0.11)}$ & $\mathbf{2.13(0.22)}$
\\

\midrule
\multirow{5}{*}{\rotatebox[origin=c]{90}{CIFAR-10}}
& Forward & $\mathbf{0.00(0.00)}$ & $47.63(0.35)$ & $39.09(0.28)$ & $41.70(0.32)$ & $35.63(0.81)$ & $45.52(0.65)$
\\
& T-Revision & $0.03(0.03)$ & $43.05(0.36)$ & $39.13(0.22)$ & $40.80(0.30)$ & $34.82(0.67)$ & $43.05(0.52)$
\\
& Dual-T & $0.81(0.04)$ & $\mathbf{2.99(0.23)}$ & $19.37(0.45)$ & $16.84(0.61)$ & $4.60(0.31)$ & $8.80(1.57)$
\\
& TVG & $0.64(0.01)$ & $3.17(0.19)$ & $\mathbf{1.56(0.13)}$ & $\mathbf{2.16(0.22)}$ & $\mathbf{1.94(0.18)}$ & $\mathbf{2.24(0.26)}$
\\
& TVD & $7.87(0.10)$ & $6.90(0.18)$ & $8.46(0.17)$ & $8.70(0.24)$ & $7.06(0.14)$ & $7.98(0.38)$
\\

\midrule
\multirow{5}{*}{\rotatebox[origin=c]{90}{CIFAR-100}}
& Forward & $\mathbf{0.00(0.00)}$ & $48.62(0.11)$ & $39.81(0.03)$ & $43.57(0.04)$ & $40.92(0.07)$ & $49.06(0.10)$
\\
& T-Revision & $0.46(0.05)$ & $31.58(0.46)$ & $39.45(0.03)$ & $42.77(0.06)$ & $40.01(0.09)$ & $39.49(0.26)$
\\
& Dual-T & $3.10(0.08)$ & $17.10(0.18)$ & $33.26(0.20)$ & $33.79(0.26)$ & $\mathbf{23.56(0.43)}$ & $22.59(0.23)$
\\
& TVG & $1.59(0.02)$ & $\mathbf{13.11(0.10)}$ & $37.79(0.30)$ & $38.83(0.34)$ & $30.80(0.51)$ & $\mathbf{16.47(0.18)}$
\\
& TVD & $21.98(0.11)$ & $26.46(0.15)$ & $\mathbf{29.47(0.26)}$ & $\mathbf{31.34(0.30)}$ & $23.86(0.22)$ & $35.37(0.30)$
\\

\bottomrule
\end{tabular}
\end{table*}

\section{Related Work}
\label{sec:related}
In addition to methods using the noise transition matrix explicitly and two-step methods detecting anchor points from noisy data \citep{patrini2017making, yu2018learning, xia2019anchor, yao2020dual} introduced in \cref{sec:introduction,sec:problem}, in this section we review related work in learning from noisy labels in a broader sense.

First, in CCN, is it possible to learn a correct classifier \emph{without} the noise transition matrix?
Existing studies in \emph{robust loss functions} \citep{ghosh2017robust, zhang2018generalized, wang2019symmetric, charoenphakdee2019symmetric, ma2020normalized, feng2020can, lyu2020curriculum, liu2020peer} showed that it is possible to alleviate the label noise issue even without estimating the noise rate/transition matrix, under various conditions such as the noise being symmetric (the RCN model in binary classification \citep{angluin1988learning}).
Further, it is proven that the accuracy metric itself can be robust \citep{chen2021robustness}.
However, if the noise is heavy and complex, robust losses may perform poorly.
This motivates us to evaluate our method under various types of label noises beyond the symmetric noise.

Another direction is to learn a classifier that is robust against label noise, including
\emph{training sample selection} 
\citep{malach2017decoupling, jiang2018mentornet, han2018co, wang2018iterative, yu2019does, wei2020combating, mirzasoleiman2020coresets, wu2020topological} that selects training examples during training,
\emph{learning with rejection}
\citep{el2010foundations, thulasidasan2019combating, mozannar2020consistent, charoenphakdee2020classification} that abstains from using confusing instances,
\emph{meta-learning}
\citep{shu2019meta, li2019learning},
and \emph{semi-supervised learning} 
\citep{nguyen2019self, li2020dividemix}.
These methods exploit the training dynamics, characteristics of loss distribution, or information of data itself instead of the class-posteriors.
Then, the CCN assumption in \cref{eq:ccn} might not be needed but accordingly these methods usually have limited consistency guarantees.
Moreover, the computational cost and model complexity of these methods could be higher.

For the CCN model and noise transition matrix estimation, recently, the idea of solving the class-conditional label noise problem using a one-step method was concurrently used by \citet{li2021provably}, aiming to relax the anchor point assumption.
They adopt a different approach based on the characteristics of the noise transition matrix, instead of the properties of the clean class-posterior used in our work.
\citet{li2021provably} has the advantage that their assumption is weaker than ours.
However, the additional term on the transition matrix might be incompatible with derivative-free optimization, such as the Dirichlet posterior update method proposed in our work.

\newpage
\section{Experiments}
\label{sec:ex}

In this section, we present experimental results to show that the proposed method achieves lower estimation error of the transition matrix and consequently better classification accuracy for the true labels, confirming \cref{thm:consistency}.


\subsection{Benchmark Datasets}

We evaluated our method on three image classification datasets, namely \textbf{MNIST} \citep{lecun1998gradient}, \textbf{CIFAR-10}, and \textbf{CIFAR-100} \citep{krizhevsky2009learning}.
We used various noise types besides the common symmetric noise and pair flipping noise.

Concretely, noise types include:
\begin{enumerate*}[label=(\alph*), itemjoin={{; }}, itemjoin*={{; and }}]
\item (\textbf{Clean})
no additional synthetic noise, which serves as a baseline for the dataset and model

\item (\textbf{Symm.}) 
symmetric noise $50\%$ \citep{patrini2017making}

\item (\textbf{Pair})
pair flipping noise $40\%$ \citep{han2018co}

\item (\textbf{Pair$^2$})
a product of two pair flipping noise matrices with noise rates $30\%$ and $20\%$.
Because the multiplication of pair flipping noise matrices is commutative, it is guaranteed to have multiple ways of decomposition of the transition matrix

\item (\textbf{Trid.}) 
tridiagonal noise \citep[see also][]{han2018masking}, which corresponds to a spectral of classes where adjacent classes are easier to be \emph{mutually} mislabeled, unlike the \emph{unidirectional} pair flipping

\item (\textbf{Rand.})
random noise constructed by sampling a Dirichlet distribution and mixing with the identity matrix to a specified noise rate
\end{enumerate*}.
See \cref{app:ex} for details.


\paragraph{Methods.}

We compared the following methods:
\begin{enumerate*}[label=(\arabic*), itemjoin={{; }}, itemjoin*={{; and }}]
\item (\textbf{MAE})
mean absolute error \citep{ghosh2017robust} as a robust loss

\item (\textbf{CCE})
categorical cross-entropy loss

\item (\textbf{GCE})
generalized cross-entropy loss \citep{zhang2018generalized}

\item (\textbf{Forward})
forward correction \citep{patrini2017making} based on anchor points detection

\item (\textbf{T-Revision})
transition-revision \citep{xia2019anchor} where the transition matrix is further revised during the second stage of training

\item (\textbf{Dual-T})
dual-T estimator \citep{yao2020dual} that uses the normalized confusion matrix to correct the transition matrix

\item (\textbf{TVG})
total variation regularization with the gradient-based estimation of $\mT$

\item (\textbf{TVD})
the one with the Dirichlet posterior update
\end{enumerate*}.


\paragraph{Models.}

For MNIST, we used a sequential convolutional neural network (CNN) and an Adam optimizer \citep{kingma2014adam}.
For both CIFAR-10 and CIFAR-100, we used a residual network model ResNet-18 \citep{he2016deep} and a stochastic gradient descent (SGD) optimizer with momentum \citep{sutskever2013importance}.


\paragraph{Hyperparameters.}

For the gradient-based estimation, we initialized the unconstrained matrix with diagonal elements of $\log(0.5)$ and off-diagonal elements of $\log(0.5 / (K-1))$, so after applying softmax the diagonal elements are $0.5$.
For the Dirichlet posterior update method, we initialized the concentration matrix with diagonal elements of $10$ for MNIST and $100$ otherwise and off-diagonal elements of $0$.
We set $\vbeta = (0.999, 0.01)$ and $\gamma = 0.1$.
We sampled $512$ (the same as the batch size) pairs in each batch to calculate the pairwise total variation distance.
Other hyperparameters are provided in \cref{app:ex}.


\paragraph{Evaluation metrics.}

In addition to the test \emph{accuracy}, we reported the \emph{average total variation} to evaluate the transition matrix estimation, which is defined as follows:
\begin{equation*}
  \frac1K \sum_{i=1}^K 
  d_\TV(\mT_i, \mTh_i)
= \frac1K \sum_{i=1}^K 
  \frac12 \sum_{j=1}^K \abs*{\mT_{ij} - \mTh_{ij}}
\in [0, 1]
.
\end{equation*}


\paragraph{Results.}

We ran $10$ trials for each experimental setting and reported ``mean (standard deviation)'' of the accuracy and average total variation in \cref{tab:accuracy,tab:tv}, respectively.
Outperforming methods are highlighted in boldface using one-tailed t-tests with a significance level of $0.05$.

In \cref{tab:accuracy}, we observed that the proposed methods performs well in terms of accuracy.
Note that a baseline method Dual-T also showed superiority in some settings, which sheds light on the benefits of using the confusion matrix.
However, as a two-step method, their computational cost is at least twice ours.
In \cref{tab:tv}, we can confirm that in most settings, our methods have lower estimation error of the transition matrix than baselines, sometimes by a large margin.
For better reusability, we fixed the initial transition matrix/concentration parameters across all different noise types.
If we have more prior knowledge about the noise, a better initialization may further improve the performance.


\begin{table}[t]
\centering
\caption{%
\textbf{Accuracy} ($\%$) on the Clothing1M dataset.
}
\label{tab:clothing1m}
\begin{tabular}{ccccc}
\toprule
 CCE & Forward & T-Revision & Dual-T & TVD
\\
\midrule
$69.91$ & $69.96$ & $69.97$ & $70.67$ & $71.65$
\\
\bottomrule
\end{tabular}
\end{table}

\begin{figure}
\centering
\includegraphics[width=\linewidth]
{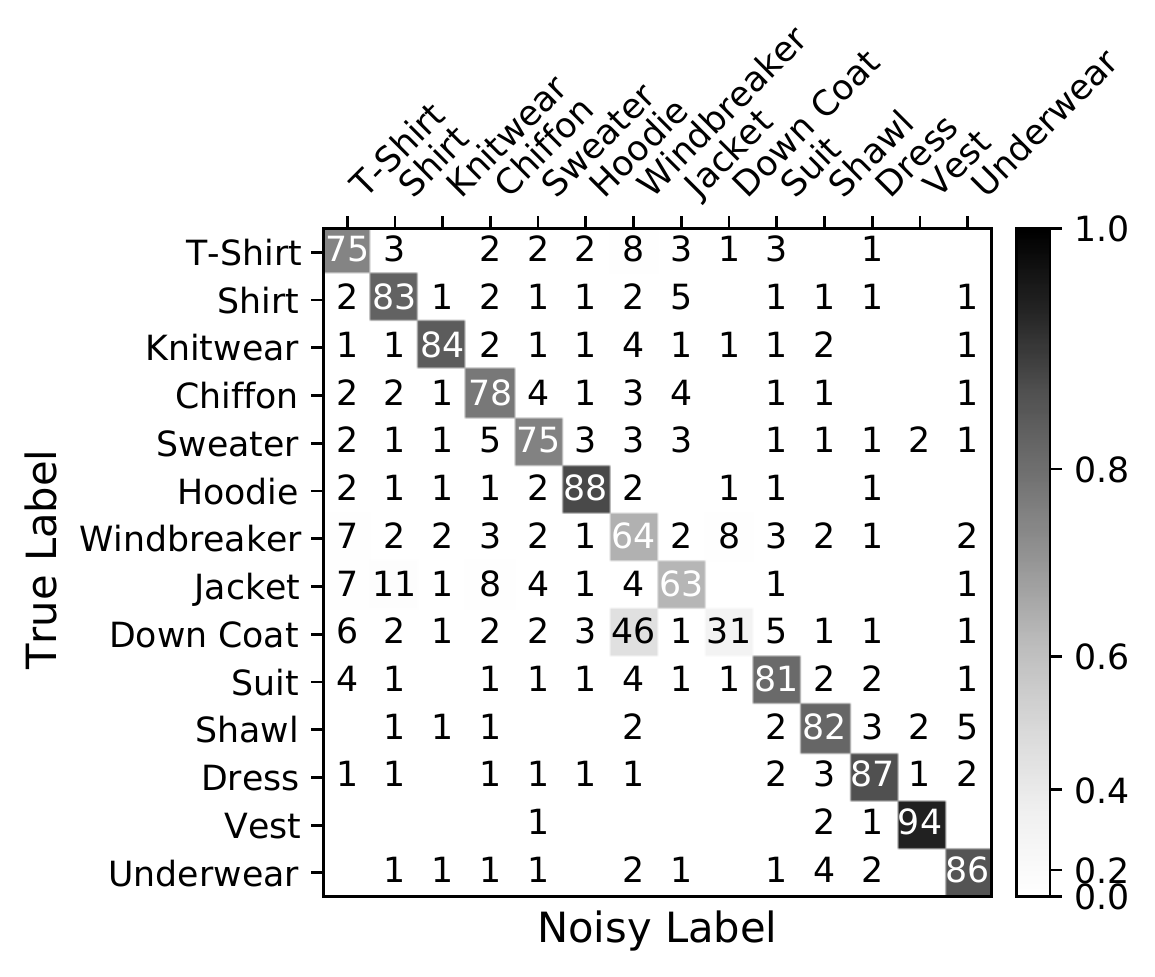}
\caption{%
\textbf{Estimated transition matrix} ($\times 100$) on Clothing1M.
}
\label{fig:T_clothing1m}
\end{figure}


\newpage
\subsection{Real-World Dataset}

We also evaluated our method on a real-world noisy label dataset, Clothing1M \citep{xiao2015learning}.
Unlike some previous work that also used a small set of clean training data \citep{patrini2017making, xia2019anchor}, we only used the $1M$ noisy training data.
We followed previous work for other settings such as the model and optimization.
We implemented data-parallel distributed training on $64$ NVIDIA Tesla P100 GPUs by PyTorch \citep{paszke2019pytorch}.
See \cref{app:ex} for details.


\paragraph{Results.}

In \cref{tab:clothing1m}, we reported the test accuracy.
The transition matrix estimated by our proposed method was plotted in \cref{fig:T_clothing1m}.

We can see that our method outperformed the baselines in terms of accuracy, which demonstrated the effectiveness of our method in real-world settings.
Although there is no ground-truth transition matrix for evaluation, we can observe the \emph{similarity relationship} between categories from the estimated transition matrix, which itself could be of great interest.
For example, if two categories are relatively easy to be mutually mislabeled, they may be visually similar; if one category can be mislabeled as another, but not vice versa, we may get a semantically meaningful hierarchy of categories.
Further investigation is left for future work.

\newpage
\section{Conclusion}
We have introduced a novel method for estimating the noise transition matrix and learning a classifier simultaneously, given only noisy data.
In this problem, the supervision is insufficient to identify the true model, i.e., we have a class of observationally equivalent models.
We address this issue by finding characteristics of the true model under realistic assumptions and introducing a partial order as a regularization.
As a result, the proposed \emph{total variation regularization} is theoretically guaranteed to find the optimal transition matrix under mild conditions, which is reflected in experimental results on benchmark datasets.

\section*{Acknowledgements}
We thank Xuefeng Li, Tongliang Liu, Nontawat Charoenphakdee, Zhenghang Cui, and Nan Lu for insightful discussion.
We also would like to thank the Supercomputing Division, Information Technology Center, the University of Tokyo, for providing the Reedbush supercomputer system.
YZ was supported by Microsoft Research Asia D-CORE program and RIKEN's Junior Research Associate (JRA) program.
GN and MS were supported by JST AIP Acceleration Research Grant Number
JPMJCR20U3, Japan.
MS was also supported by the Institute for AI and Beyond, UTokyo.

\bibliography{references}


\clearpage
\appendix
\onecolumn
\everymath{\displaystyle}


\section{Beyond Class-Conditional Noise}
\label{app:ccn}

In this section, we provide an overview of several selected noise models.

As introduced in \cref{sec:introduction}, early studies focused on the most simple case --- the \emph{random classification noise} (RCN) model for binary classification \citep{angluin1988learning, long2010random, van2015learning}, where binary labels are flipped independently with a fixed noise rate $\rho \in [0, 0.5)$.
Here, $Y, \Yt \in \{\pm1\}$ and $\rho = p(\Yt = -1 | Y = +1) = p(\Yt = -1 | Y = +1)$.
Then, still for binary classification, the \emph{class-conditional noise} (CCN) model \citep{natarajan2013learning} extended the RCN model to the case where the noise rate depends on the class: $\rho_{+1} = p(\Yt = -1 | Y = +1)$, $\rho_{-1} = p(\Yt = +1 | Y = -1)$, $\rho_{+1} + \rho_{-1} < 1$.
These noise models are special cases of the multiclass CCN model \citep{patrini2017making, goldberger2017training, han2018masking, xia2019anchor, yao2020dual}, which is the main focus of our work.

A more general framework for learning with label noise is the \emph{mutual contamination} (MC) model
\citep{scott2013classification, blanchard2014decontamination, du2014analysis, menon2015learning, lu2019on}, where examples of each class are drawn separately.
That is, $p(X | Y)$ is corrupted but not $p(Y | X)$.
Consequently, the marginal distribution of data may not match the true marginal distribution.
It is known that CCN is a special case of MC \citep{menon2015learning}.
For the binary case, there is a related problem of the transition matrix estimation, called \emph{mixture proportion estimation} (MPE) \citep{du2014analysis, scott2015rate, ramaswamy2016mixture}, which has more technical difficulties.
Our method may not work well in the MC setting because it explicitly relies on the i.i.d.~assumption.

Further, the \emph{instance-dependent noise} (IDN) model \citep{menon2018learning, cheng2020learning, berthon2021confidence} has been assessed to only a limited extent but is of great interest recently.
IDN still explicitly models the label corruption process as CCN bu removes the CCN assumption in \cref{eq:ccn}.
Therefore, the noise transition matrix could be instance-dependent and thus harder to estimate.
In other words, $\mT$ is not a fixed matrix anymore but a matrix-valued function of the instance $\mT(X): \sX \to \sT$.
Owing to its complexity, IDN has not been investigated extensively.

One simple way is to estimate the matrix-valued function $\mT(X)$ and the clean class-posterior $p(Y | X)$ directly, assuming a certain level of smoothness of $\mT(X)$ \citep{goldberger2017training}.
However, there is no theoretical guarantee and the estimation error could be very high.
Another direction is to restrict the problem so we could provide some theoretical guarantees under certain conditions \citep{menon2018learning, cheng2020learning}.
It is also a promising way to approximate IDN using a simpler dependency structure \citep{xiao2015learning, xia2020part}, which works well in practice.
Our method may also serve as a practical approximation of IDN without theoretical guarantee, which is reflected in the experiment on the Clothing1M dataset (\cref{sec:ex}).
The use of regularization techniques in our work may inspire practical algorithm design for IDN.

Learning from noisy labels has a rich literature and there are several other noise models, e.g., capturing the uncertainty of labels without explicitly modeling the label corruption process.
Although they are out of scope of our discussion.


\clearpage
\section{Overconfidence in Neural Networks}
\label{app:overconfidence}

\begin{figure}
\centering
\includegraphics[width=\linewidth]
{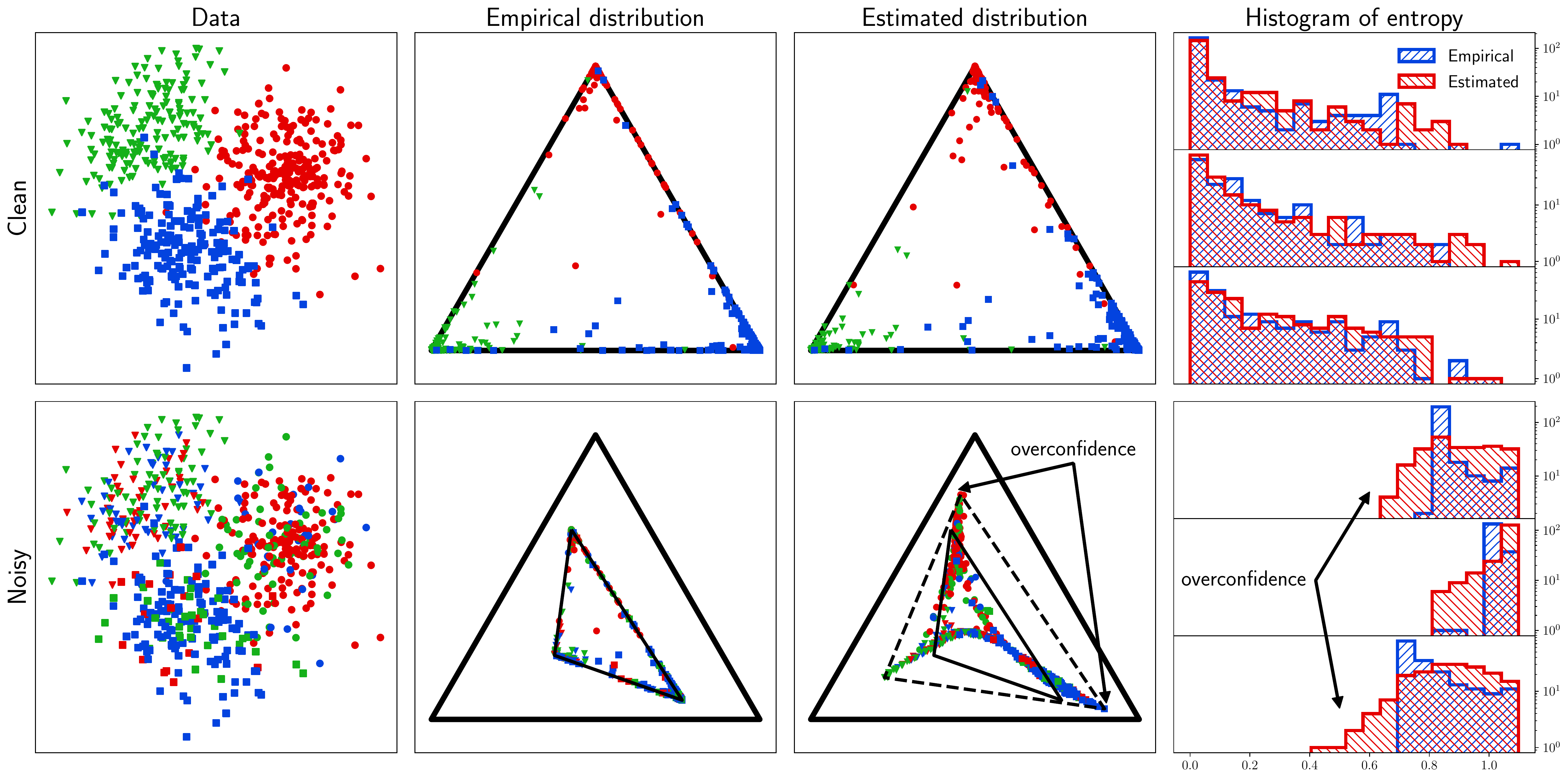}
\vspace{-2.5em}
\caption[]{%
An example of \textbf{overconfident predictions} yield from neural networks, comparing the clean/noisy class-posterior estimation.
See \cref{fig:overconfidence} for the notation.
}
\label{fig:overconfidence_comparison}
\end{figure}

In this section, we discuss the overconfidence phenomenon in neural networks, which is partially presented in \cref{ssec:unknown_T}.

In neural network training, if we only care the classification accuracy, the overconfidence is not a problem.
We could use any \emph{classification-calibrated} loss \citep{bartlett2006convexity, tewari2007consistency}, which only guarantees that the accuracy (0-1 risk) is asymptotically optimal.
The class-posterior might not be recovered from the output of the classifier.

However, the transition matrix estimation based on anchor points \citep{patrini2017making, xia2019anchor} --- the method shown in \cref{ssec:unknown_T} --- heavily relies on a \emph{confidence-calibrated} estimation of the noisy class-posterior using neural networks.
The overconfidence issue of possibly over-parameterized neural networks has been discovered, and several re-calibration methods are developed to alleviate this issue \citep{guo2017calibration, kull2019beyond, hein2019relu, rahimi2020intra}.

In learning with class-conditional label noise, we can demonstrate that estimating the noise class-posterior causes a significantly worse overconfidence issue than estimating the clean class-posterior.
\cref{fig:overconfidence_comparison} shows a comparison between the clean class-posterior estimation and the noisy class-posterior estimation (also shown in \cref{fig:overconfidence}).
We used a Gaussian mixture with $3$ components as the training data, a $3$-layer multilayer perceptron (MLP) with hidden layer size of $32$ and rectified linear unit (ReLU) activations as the model, and an Adam optimizer \citep{kingma2014adam} with batch size of $64$ and learning rate of $\num{1e-3}$.

As discussed in \cref{ssec:ccn}, $\vp(\Yt | X)$ should be within the convex hull $\Conv(\mT)$.
However, without knowing $\mT$ and this constraint, a neural network trained with noisy labels tends to output overconfident probabilities that are outside of $\Conv(\mT)$.
The lack of the constraint $\Conv(\mT)$ aggravates the overconfidence problem and might make it harder to re-calibrate the confidence.
Consequently, transition matrix estimation may suffer from poorly estimated noisy class-posteriors, which leads to performance degradation of the aforementioned two-step methods.

This is the motivation of using the product $\mTh\T\vph(Y | X)$ as an estimate of $\vp(\Yt | X)$ and avoiding estimating $\vp(\Yt | X)$ directly using neural networks.
However, it is important to note that the neural network may still suffer from the overconfidence issue, especially after we enforce the predicted probabilities to be more distinguishable from each other in terms of the pairwise total variation distance.
In such a case, if the confidence of $\vp(Y | X)$ is needed to make decisions, post-hoc re-calibration methods can be applied \citep{guo2017calibration, kull2019beyond, hein2019relu, rahimi2020intra}.
Nevertheless, if we only use the accuracy as the evaluation metric, the overconfidence issue in the clean class-posterior estimation is much less harmful than it in the noisy class-posterior estimation, which affects the transition matrix estimation significantly.


\clearpage
\section{Proof}
\label{app:proof}

In this section, we provide the proof of \cref{thm:consistency}:
\consistency*

First, recall \cref{thm:decomposition}:
\decomposition*

and the definitions of $L_0(W, \mTh)$ and $R(W)$:
\begin{align}
\tag{\ref{eq:kl}}
  L_0(W, \mTh)
&\defeq
  \E_{X \sim p(X)}
  \brackets*{
  D_\KL \diver*{\vp(\Yt | X)}{\mTh\T\vph(Y | X; W)}
  }
,
\\
\tag{\ref{eq:regularization}}
  R(W) 
&\defeq 
  \E_{x_1 \sim p(X)}\E_{x_2 \sim p(X)}
  [d_\TV(\vph_1, \vph_2)]
,
\where
  \vph_i \defeq \vph(Y | X = x_i; W), \quad i = 1, 2
.
\end{align}
Also, recall that we considered a sufficiently large function class of $\vph(Y | X; W)$ that contains the ground-truth $\vp(Y | X)$, i.e.,
$
\exists W^* \in \sW, 
\vph(Y | X; W^*) = \vp(Y | X) 
\almosteverywhere
$
Although there could be a set of $W^*$ satisfying this condition, without loss of generality, we assume that $W^*$ is unique.

Denote the set of $W$ and $\mTh$ s.t.~$L_0(W, \mTh) = 0$ by $(\sW \times \sT)_0 \subset \sW \times \sT$.
By definition, $(W^*, \mT) \in (\sW \times \sT)_0$.

Then, we have the following lemmas:
\begin{lemma}
\label{lem:identity_of_indiscernibles}
$\forall (W, \mTh) \in (\sW \times \sT)_0$,
$\exists \mU \in \sT$,
$
  \vp(\Yt | X)
= \mT\T \vp(Y | X)
= \mTh\T \vph(Y | X; W)
= \mTh\T (\mU\T \vp(Y | X))
\almosteverywhere
$
\vspace{-1em}
\begin{proof}
This is the due to the \emph{identity of indiscernibles} property of the KL-divergence and \cref{thm:decomposition}.
\end{proof}
\end{lemma}

\begin{lemma}
\label{lem:optimal_parameter}
$\forall (W, \mTh) \in (\sW \times \sT)_0$,
$R(W) \leq R(W^*)$.
\vspace{-1em}
\begin{proof}
This is a direct consequence of \cref{lem:identity_of_indiscernibles}, the contraction \cref{eq:contraction}, and our assumption of the existence of $W^*$:
\begin{align}
  R(W)
&= 
  \E_{x_1 \sim p(X)}\E_{x_2 \sim p(X)}
  [d_\TV(\vph(Y | X = x_1; W), \vph(Y | X = x_2; W))]
\\
&= 
  \E_{x_1 \sim p(X)}\E_{x_2 \sim p(X)}
  [d_\TV(\mU\T \vp(Y | X = x_1), \mU\T \vp(Y | X = x_2))]
\\
&\leq
  \E_{x_1 \sim p(X)}\E_{x_2 \sim p(X)}
  [d_\TV(\vp(Y | X = x_1), \vp(Y | X = x_2))]
\\
&= 
  \E_{x_1 \sim p(X)}\E_{x_2 \sim p(X)}
  [d_\TV(\vph(Y | X = x_1; W^*), \vph(Y | X = x_2; W^*))]
=
  R(W^*)
.
\end{align}
\end{proof}
\end{lemma}

\begin{lemma}
\label{lem:uniform_convergence}
$\sup_{W \in \sW} 
\abs*{R(W) - \Rt(W)} 
\convergep 0$ 
and
$\sup_{\substack{W \in \sW \\ \mTh \in \sT}} 
\abs*{L_0(W, \mTh) - \Lt_0(W, \mTh)} 
\convergep 0$
as $N \to \infty$.
\vspace{-1em}
\begin{proof}
This is due to the i.i.d.~assumption, the compactness of $\sW$ and $\sT$, the continuity of $R(W)$ and $L_0(W, \mTh)$, and the \emph{uniform law of large numbers}.
\end{proof}
\end{lemma}

Finally, by the definition and \cref{lem:uniform_convergence}, 
we have
$\mathbb{P}\brackets*{L_0(W^\circ, \mTh^\circ) = 0} \to 1$, 
and by \cref{lem:optimal_parameter,lem:uniform_convergence}, 
we have 
$W^\circ \to W^*$ as $N \to \infty$.
Therefore, $\vph(Y | X; W^\circ) \converged \vp(Y | X) \almosteverywhere$, which means that the corresponding $\mU \converge \mI$ and thus $\mTh^\circ \converge \mT$ as $N \to \infty$. 
\hfill$\blacksquare$


\clearpage
\section{Intuition}

\begin{figure}
\centering
\includegraphics[width=\linewidth]
{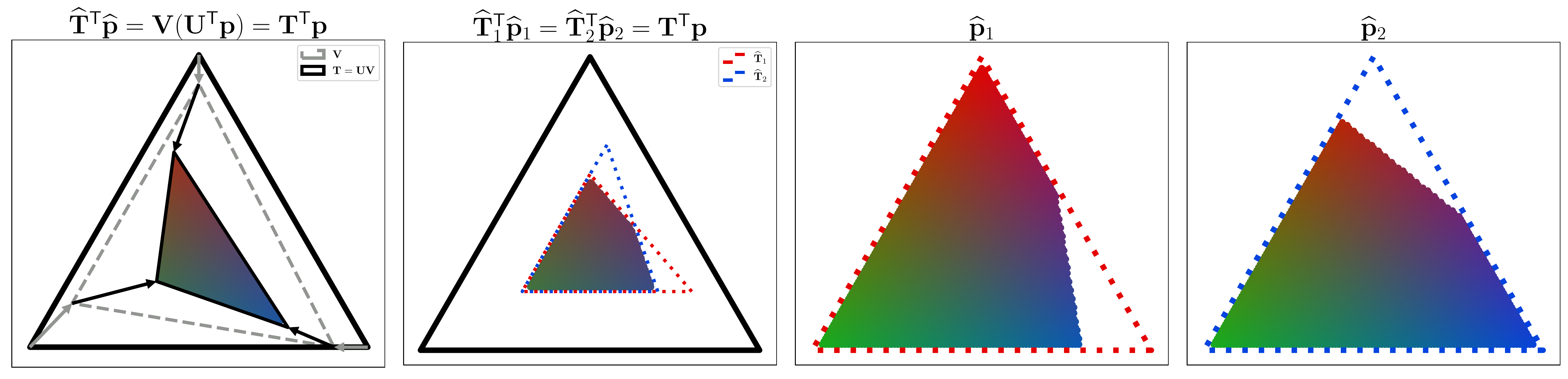}
\vspace{-2em}
\caption{%
The intuition behind the theoretical results.
}
\label{fig:intuition}
\end{figure}

In this section, we discuss the intuition behind our theoretical results.

Given the only constraint of $\vp(\Yt | X) = \mTh\T\vph(Y | X)$, a feasible $\mTh$ can be any matrix that satisfies
\begin{equation}
\forall x \in \sX, 
\vp(\Yt | X=x) \in \Conv(\mTh)
,
\end{equation}
and we can find $\vph(Y | X)$ accordingly, if the function class is sufficiently large.
This is the partial identifiability problem.
On the other hand, in real-world problems, we hope that the clean class-posterior $\vp(Y | X)$ is the ``cleannest'', so at least 
\begin{equation}
\nexists \mS \in \sT,
\st \vp(Y | X) = \mS\T\vp'(Y | X)
.
\end{equation}
Otherwise, $\vp(\Yt | X) = (\mS\mT)\T \vp'(Y | X)$ also holds and $\vp'(Y | X)$ might be a better solution.
Thus, we become less ambitious, ignore all intermediate possible solutions and only aim to find the ``cleannest'' one.

However, there could be multiple ``cleannest'' ones in this sense.
An example is illustrated in \cref{fig:intuition}.
There could be $\mTh_1$, $\mTh_2$, and $\vph_1(Y | X)$, $\vph_2(Y | X)$, such that 
\begin{equation}
\vp(\Yt | X) = \mTh_1\T\vph_1(Y | X) = \mTh_2\T\vph_2(Y | X)
,
\end{equation}
and both $\vph_1(Y | X)$ and $\vph_2(Y | X)$ satisfy the condition above.
In this sense, we still cannot distinguish $\vph_1(Y | X)$ and $\vph_2(Y | X)$.
Either of them can be the true clean class-posterior.

To avoid such cases, in this work, we made the assumption that anchor points exist, i.e., there are instances for each class that we are absolutely sure which class they belong to.
Such instances are considered prototypes of each class, and we believe that they exist in many real-world noisy datasets.
In this way, we can guarantee the uniqueness of the ``cleannest'' clean class-posterior and the transition matrix, and consequently construct consistent estimators to find them, as explained in this paper.

If anchor points for all classes do not exist, the proposed algorithm may still work in practice but there is no theoretical guarantee yet.
As mentioned in \cref{sec:related}, \citet{li2021provably} aims to relax the anchor point assumption.
From the perspective of the geometric property of the transition matrix, it is possible to solve this problem in a weaker condition, which is, however, not the focus of this work.


\clearpage
\section{Experiments}
\label{app:ex}

In this section, we provide missing details of the experimental settings used in \cref{sec:ex}.


\subsection{Benchmark Datasets}

\begin{figure}
\centering
\includegraphics[width=.8\linewidth]
{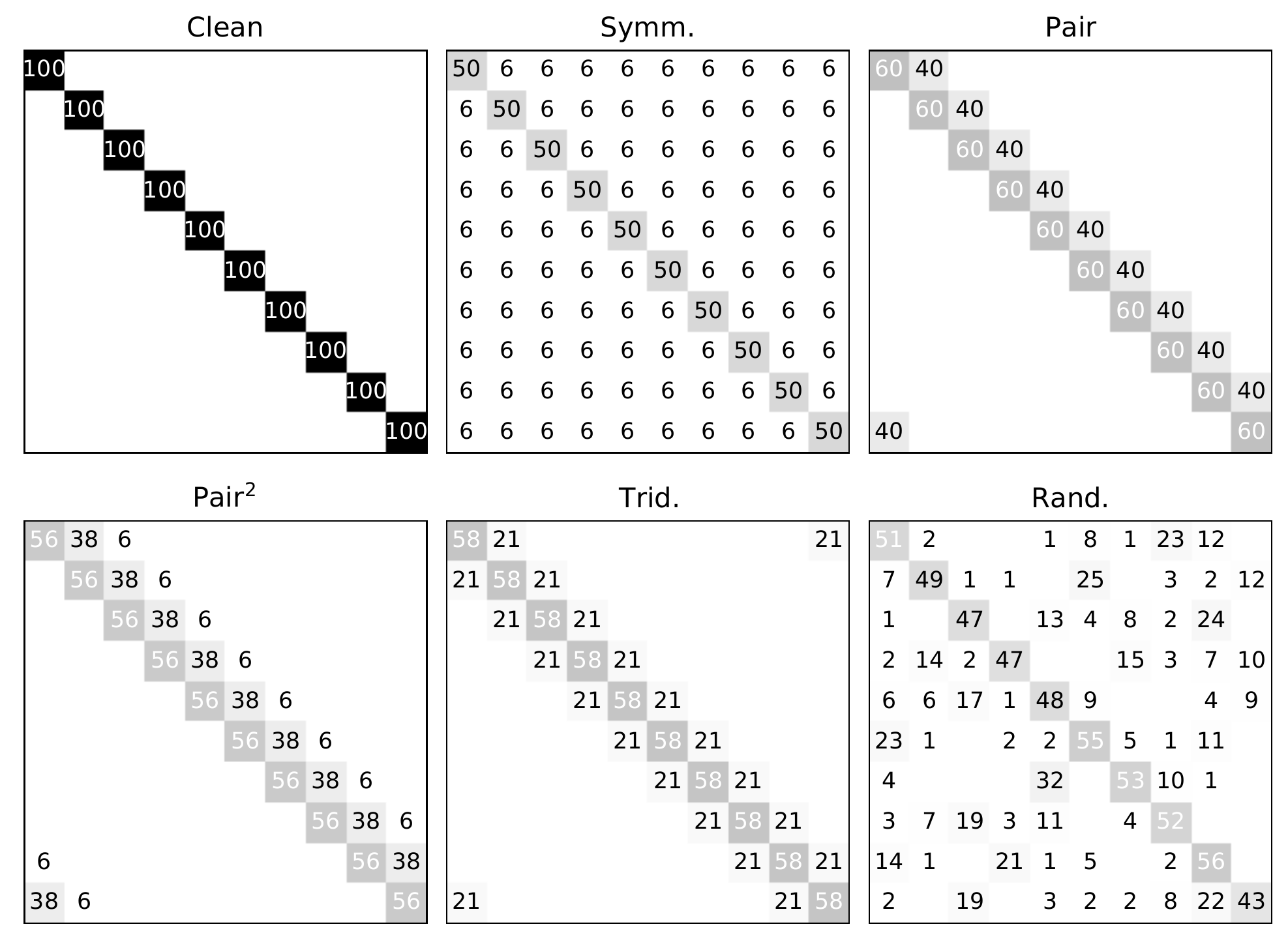}
\vspace{-1em}
\caption{%
\textbf{Synthetic transition matrices} ($\times 100$) used in our experiments when $K = 10$ (MNIST and CIFAR-10).
The random noise matrix (Rand.) is an example while other matrices are fixed.
}
\label{fig:T_synthetic}
\end{figure}

\paragraph{Data.}

We used the MNIST,\footnote{%
MNIST \citep{lecun1998gradient}
\url{http://yann.lecun.com/exdb/mnist/}
}
CIFAR-10, and CIFAR-100\footnote{%
CIFAR-10, CIFAR-100 \citep{krizhevsky2009learning}
\url{https://www.cs.toronto.edu/~kriz/cifar.html}
}
datasets.
The MNIST dataset contains $28 \times 28$ grayscale images in $10$ classes.
The size of the training set is $60000$ and the size of the test set is $10000$.
The CIFAR-10 and CIFAR-100 datasets contain $32 \times 32$ colour images in $10$ classes and in $100$ classes, respectively.
The size of the training set is $50000$ and the size of the test set is $10000$.

\paragraph{Data preprocessing.}

For MNIST, We did not use any data augmentation.
For CIFAR-10 and CIFAR-100, we used random crop and random horizontal flip.
We added synthetic label noise into the training sets.
The test sets were not modified.
Overall, the transition matrices are plotted in \cref{fig:T_synthetic}.
More specifically, we used:
\begin{enumerate}
\item (\textbf{Clean})
no additional synthetic noise, which serves as a baseline for the dataset and model.

\item (\textbf{Symm.}) 
symmetric noise with noise rate $50\%$ \citep{patrini2017making}.

\item (\textbf{Pair})
pair flipping noise with noise rate $40\%$ \citep{han2018co}.

\item (\textbf{Pair$^2$})
a product of two pair flipping noise matrices with noise rates $30\%$ and $20\%$.
Because the multiplication of pair flipping noise matrices is commutative, it is guaranteed to have multiple ways of decomposition of the transition matrix, e.g., $\mT_\mathrm{pair}(30\%) \mT_\mathrm{pair}(20\%) 
=\mT_\mathrm{pair}(20\%) \mT_\mathrm{pair}(30\%)$.
The overall noise rate is $44\%$.

\item (\textbf{Trid.}) 
tridiagonal noise \citep[see also][]{han2018masking}, which corresponds to a spectral of classes where adjacent classes are easier to be \emph{mutually} mislabeled, unlike the \emph{unidirectional} pair flipping.
It can be implemented by two consecutive pair flipping transformations in the opposite direction.
We used $\mT_\mathrm{pair}(30\%) \mT_\mathrm{pair}(30\%)\T$ in the experiment.
The overall noise rate is $42\%$.
Strictly, the matrix is not a tridiagonal matrix in the conventional sense because $\mT_{1,K}$ and $\mT_{K,1}$ are non-zero.

\item (\textbf{Rand.})
random noise constructed by sampling a Dirichlet distribution and mixing with the identity matrix to a specified noise rate.
The higher the concentration parameter of the Dirichlet distribution is, the more uniform the off-diagonal elements of the transition matrix are.
We used $0.5$ in the experiment.
Then, we mixed the sampled matrix with the identity matrix linearly to make the overall noise rate $50\%$.
The transition matrix is sampled for each trial.
\end{enumerate}

\paragraph{Models.}

For MNIST, we used a sequential convolutional neural network with the following structure:
\texttt{Conv2d}(channel=$32$) $\times2$,
\texttt{Conv2d}(channel=$64$) $\times2$,
\texttt{MaxPool2d}(size=$2$),
\texttt{Linear}(dim=$128$),
\texttt{Dropout}(p=$0.5$),
\texttt{Linear}(dim=$10$).
The kernel size of convolutional layers is $3$, and rectified linear unit (ReLU) is applied after the convolutional layers and linear layers except the last one.
For both CIFAR-10 and CIFAR-100, we used a ResNet-18 model \citep{he2016deep}.

\paragraph{Optimization.}

For MNIST, we used an Adam optimizer \citep{kingma2014adam} with batch size of $512$ and learning rate of $\num{1e-3}$.
The model was trained for $2000$ iterations ($17.07$ epochs) and the learning rate decayed exponentially to $\num{1e-4}$.
For CIFAR-10 and CIFAR-100, we used a stochastic gradient descent (SGD) optimizer with batch size of $512$, momentum of $0.9$, and weight decay of $\num{1e-4}$.
The learning rate increased from $0$ to $0.1$ linearly for $400$ iterations and decreased to $0$ linearly for $3600$ iterations ($4000$ iterations/$40.96$ epochs in total).

For the gradient-based estimation, we used an Adam optimizer \citep{kingma2014adam}.
The learning rate increased from $0$ to $\num{5e-3}$ linearly for $400$ iterations and creased to $0$ linearly for the rest iterations.
This is helpful because at earlier stage, the model was not sufficiently trained yet and changing the transition matrix too much may destabilize the training of the model.

We tuned hyperparameters using grid search on a small experiment and fixed them in all experimental settings.
For better reusability, we assumed that we are noise-agnostic and did not fine-tune hyperparameters for each noise type.
If we have more prior knowledge about the noise, a better initialization may further improve the performance.

\paragraph{Infrastructure.}

The experiments were conducted on NVIDIA Tesla P100 GPUs.
We used a single GPU for MNIST and data-parallel on $2$ GPUs for CIFAR-10 and CIFAR-100.

\paragraph{Results.}

\begin{figure}
\centering
\includegraphics[width=\linewidth]
{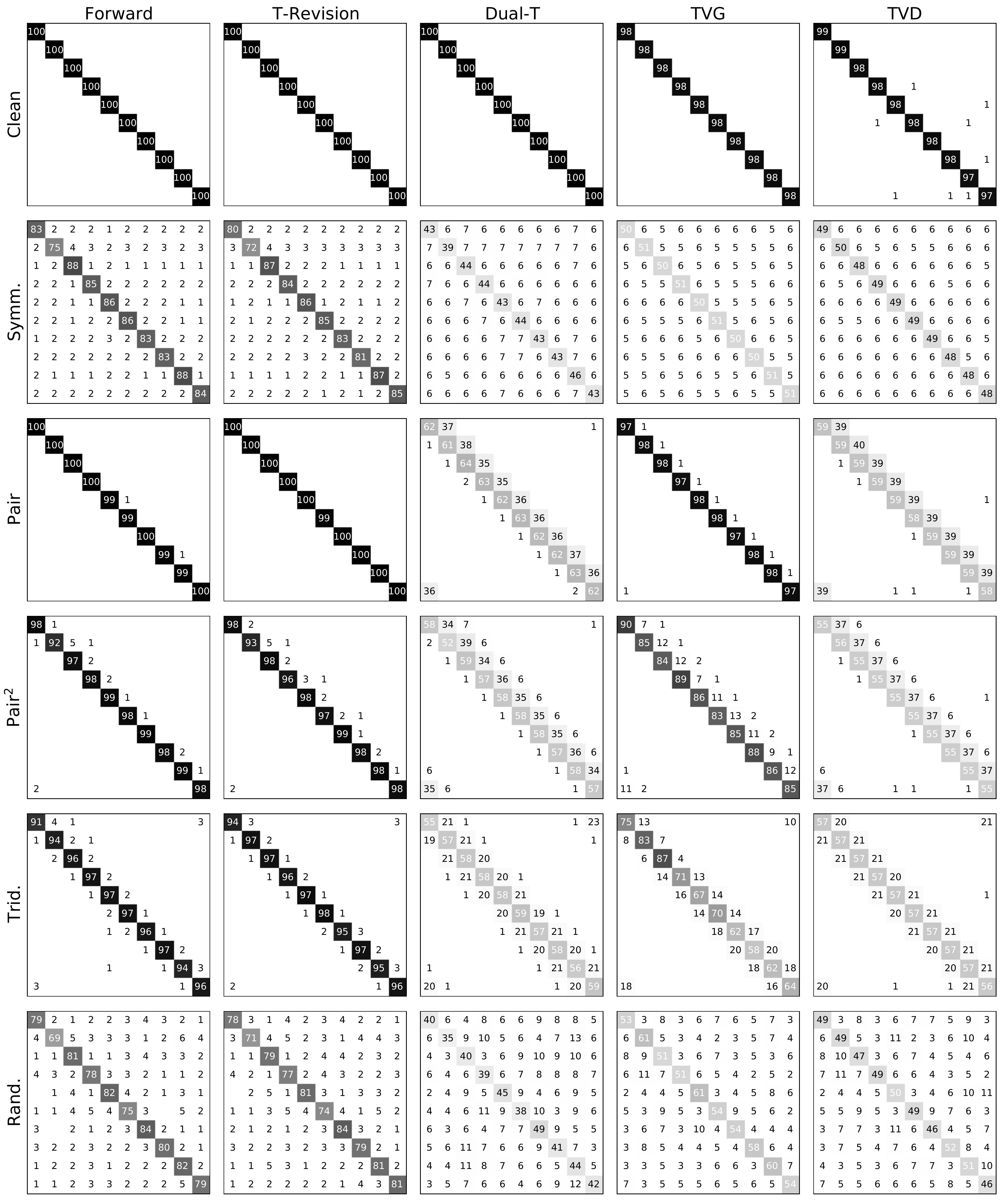}
\vspace{-2em}
\caption{%
\textbf{Estimated transition matrices} ($\times 100$) on MNIST.
}
\label{fig:T_mnist}
\end{figure}

\begin{figure}
\centering
\includegraphics[width=\linewidth]
{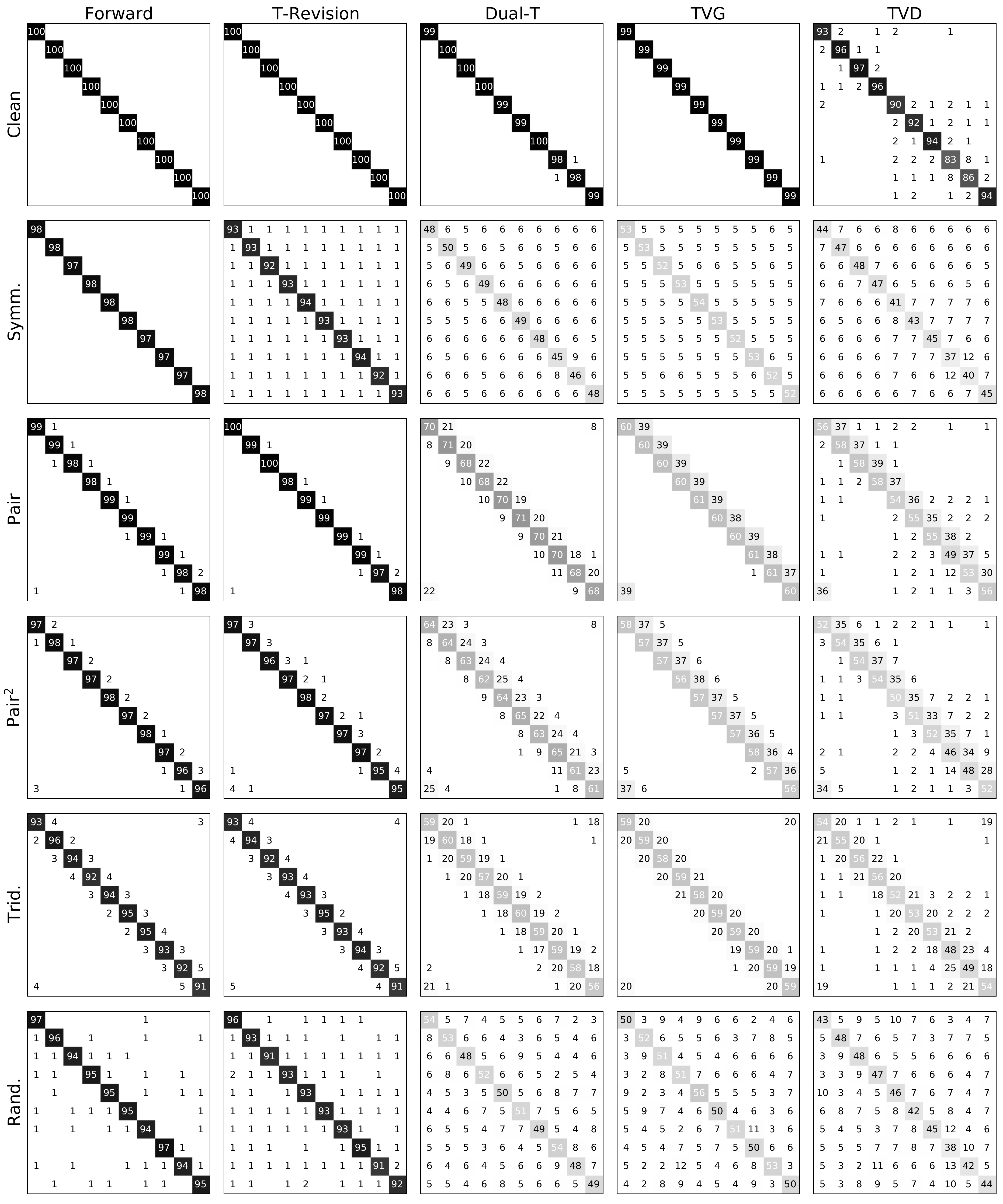}
\vspace{-2em}
\caption{%
\textbf{Estimated transition matrices} ($\times 100$) on CIFAR10.
}
\label{fig:T_cifar10}
\end{figure}

\begin{figure}
\centering
\includegraphics[width=\linewidth]
{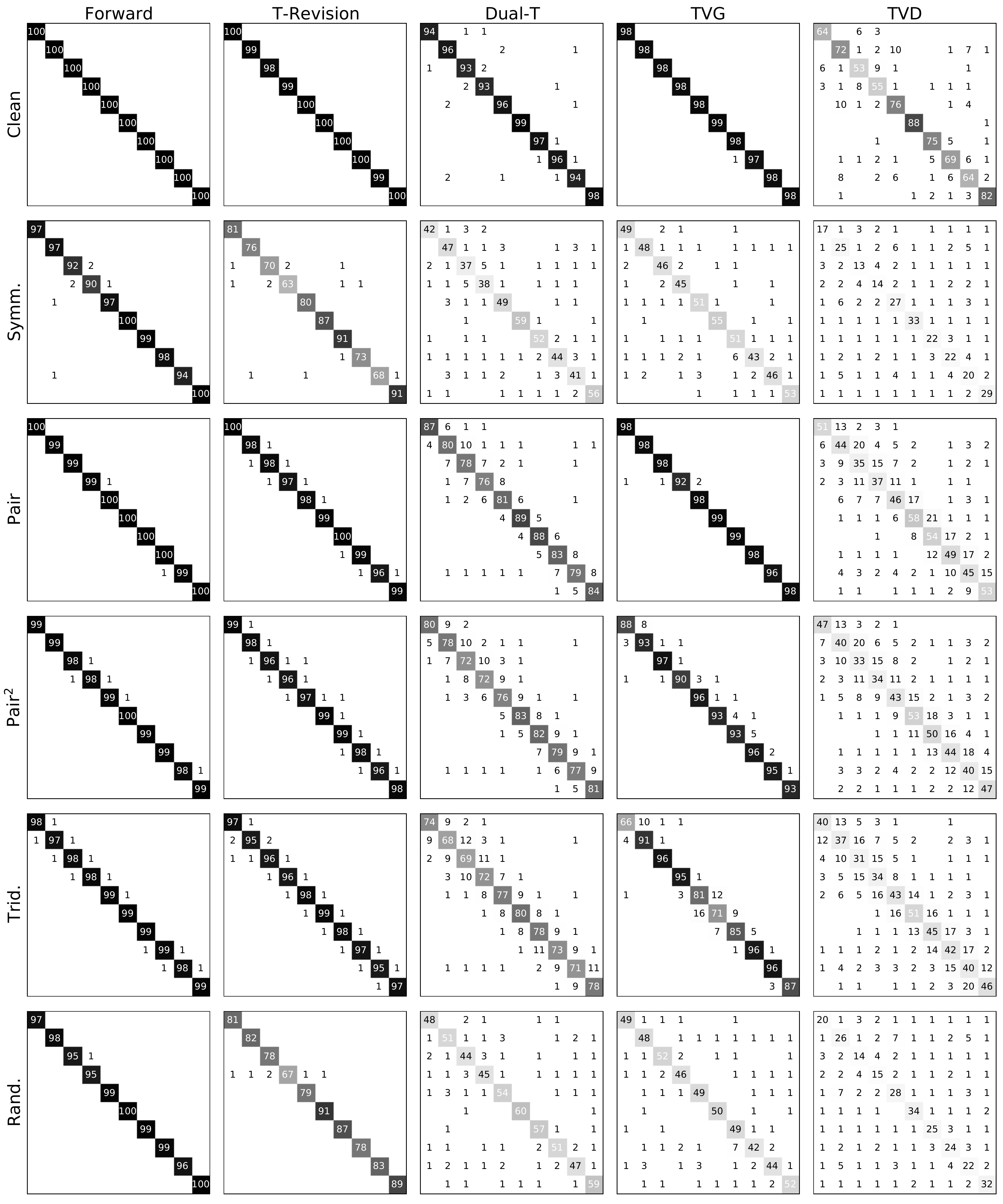}
\vspace{-2em}
\caption{%
\textbf{Estimated transition matrices} ($\times 100$) on CIFAR100 (first $10$ classes).
}
\label{fig:T_cifar100}
\end{figure}

In addition to the accuracy and average total variation presented in \cref{tab:accuracy,tab:tv}, we also provide the heat maps of the estimated transition matrices in \cref{fig:T_mnist,fig:T_cifar10,fig:T_cifar100}.
Extremely small numbers are hided for better demonstration.

We can observe that our proposed method, especially the Dirichlet posterior update method, usually has better estimation of the transition matrix under various noise types.
Dual-T \citep{yao2020dual} also performs well in some settings, which is also reflected in \cref{tab:tv}.


\subsection{Clothing1M}

\paragraph{Data.}

Clothing1M \citep{xiao2015learning} is a real-world noisy label dataset.
It contains $47570$ clean training images, $\num{1e6}$ ($1M$) noisy training images, $14313$ clean validation images, and $10526$ clean test images in $14$ classes.
We only used the noisy training data and clean test data.

\paragraph{Model and optimization.}

We followed previous work \citep{patrini2017making, xia2019anchor}.
We used a ResNet-50 model \citep{he2016deep} pretrained on ImageNet and a SGD optimizer with momentum of $0.9$, weight decay of $\num{1e-3}$, and batch size of $32$.
We trained the model on $64$ GPUs for $5000$ iterations ($10.24$ epochs in total).
The learning rate was $\num{1e-3}$ for the first half and $\num{1e-4}$ for the second half.

\paragraph{Other hyperparameters.}

We initialized the concentration matrix with diagonal elements of $1$ and off-diagonal elements of $0$.
We set $\vbeta = (0.999, 0.01)$ and $\gamma = 0.1$.

\paragraph{Infrastructure.}

We implemented data-parallel distributed training on $64$ NVIDIA Tesla P100 GPUs by PyTorch \citep{paszke2019pytorch}.
The average runtime is about $15$ (without \texttt{SyncBatchNorm}) to $25$ (with \texttt{SyncBatchNorm}) minutes.


\clearpage

\clearpage

\clearpage

\end{document}